\documentclass[10pt,twocolumn,letterpaper]{article}

\usepackage[pagenumbers]{cvpr} 

\usepackage{booktabs}
\usepackage{array}
\usepackage{colortbl}
\usepackage{multirow}
\usepackage{xspace}
\usepackage{amsmath}
\usepackage{amssymb}
\usepackage{graphicx}
\usepackage{microtype}

\definecolor{accbg}{RGB}{232,245,243}    
\definecolor{confbg}{RGB}{253,236,230}   
\definecolor{accsig}{RGB}{128,204,194}   
\definecolor{confsig}{RGB}{246,166,142}  
\definecolor{ecebg}{RGB}{236,240,250}    
\definecolor{brierbg}{RGB}{252,243,224}  
\definecolor{aurocbg}{RGB}{236,247,232}  

\newcommand{\webshop}{WebShop\xspace}
\newcommand{\sqid}{SQID\xspace}
\newcommand{\wands}{WANDS\xspace}

\definecolor{cvprblue}{rgb}{0.21,0.49,0.74}
\usepackage[pagebackref,breaklinks,colorlinks,allcolors=cvprblue]{hyperref}

\def\confName{CVPR}
\def\confYear{2027}

\renewcommand*{\paragraph}[1]{\noindent\textbf{#1}}

\title{Thinking Harder Doesn't Lead to Better Choices: Reasoning Effort Increases the Confidence but Not the Correctness of LLM Shopping Agents}

\author{Adi Krishnamoorthy\\
University of Chicago\\
{\tt\small akrishn6@chicagobooth.edu}
\and
X.Y. Han\\
University of Chicago\\
{\tt\small XY.Han@chicagobooth.edu}
}

\begin{document}
\maketitle

\begin{abstract}
Large language models (LLMs) are increasingly deployed as autonomous shopping agents that process multimodal product listings (consisting of visual and text data) and recommend purchases on behalf of consumers. Naturally, one might assume that models with more thinking power would make higher-quality purchasing decisions. To test this belief, we ask two questions about such agents: does asking a model to think harder make it choose better products, and does the confidence it reports tell us the likelihood it may be wrong? We benchmark ten frontier models on three product-choice benchmarks built from WebShop, Amazon's SQID and Wayfair's WANDS platforms to see if LLMs with more thinking power actually make better-quality product purchase decisions. Across all three benchmarks, we found that increasing the thinking effort does not increase decision quality in a statistically meaningful way. The same change in thinking effort raises stated confidence at a statistically significant level, so extra thinking buys confidence rather than correctness. We also measure the calibration of these models (how well their reported confidence percentage matches their true correctness percentage) in these tasks and show that most models are not well-calibrated at every effort level. We also find that errors are strongly shared: when one model errs in a task, several others tend to err on that same task. Furthermore, on tasks where several models all err, a large number of them pick the same wrong product.
\end{abstract}

\section{Introduction}
\label{sec:intro}

Shopping is becoming an increasingly agentic task. Assistants built on large language models (LLMs) now search catalogs, compare listings and place orders with little human review~\cite{allouah2025agentic,elhousni2026aimediation,yao2022webshop}. Two properties of such an agent matter to the person it acts for: whether its choice is preferable, and whether it is overconfident in its decision. For the latter, if an agent is overconfident, it may quietly buy the wrong thing, while if it is well-calibrated (by which we mean a model's self-reported confidence is close to its true reliability), it can hand the decision back to the user when it is unsure.

Modern frontier LLMs with which these agents are implemented now possess an explicit \emph{reasoning effort} control. Reasoning effort allows users to request low, medium or high effort. At higher reasoning efforts, a model spends correspondingly more tokens ``thinking'' before it answers~\cite{openai2024o1,deepseek2025r1}. On math and coding benchmarks this increase in ``inference-time compute'' is shown to improve accuracy~\cite{lightman2024verify,snell2024scaling}. However, whether inference-time scaling using longer thinking improves everyday decisions like shopping --- whose difficulty lies in reading a listing carefully rather than in multi-step derivation --- is less clear~\cite{sprague2024cot,chen2024overthinking}, and whether it improves calibration is largely untested.

Thus, we will study the effects of increased reasoning effort in agentic shopping using three datasets: (I) The \webshop~\cite{yao2022webshop} environment, which contains Amazon product listings with images of the product and text containing the product data; (II) the Amazon Shopping Queries dataset with \sqid images~\cite{reddy2022esci,alghossein2024sqid}, a multimodal dataset with image information and visual embeddings for different Amazon products along with associated search queries in text form; and (III) the Wayfair \wands~\cite{chen2022wands} dataset, a text-only benchmark where each product contains a title, some features, and a description. As we will describe later (\cref{sec:method}), in our experiments, we will also isolate the decision-making part of the shopping task from the search and navigation part. 

From our experiments, we first find in \cref{sec:effort} that moving from low to high effort changes {\it correctness} by only a small amount for any LLM: Only three models improve their success rate by a statistically significant amount in \webshop, and even fewer models show the same improvements in the other two datasets. However, we find that the same change in effort raises stated {\it confidence} significantly for nine of ten models in \webshop, five of ten models in \sqid, and six in \wands, so raising reasoning level delivers more confidence without corresponding correctness. Then, we find in \cref{sec:calibration} that calibration is almost entirely model-dependent: there doesn’t seem to be any correlation between calibration metrics like ECE, Brier scores, or AUROC (defined later in \cref{sec:metrics}) across models regardless of reasoning effort.
Finally, as we will show in \cref{sec:concentration}, mistakes are correlated across different models: the number of models wrong on any given task is much more concentrated than what we would expect from an independent-models baseline. Moreover, when several models err on the same task, they typically choose the same incorrect product.

\noindent\textbf{Main Contributions.} Thus, this paper makes three contributions: (I) We show that reasoning effort is largely ineffective for improving shopping decisions but does tend to make models more confident (\cref{sec:effort}). (II) We show some models are consistently much more confident, but not more accurate, than other models (\cref{sec:calibration}). (III) We provide a cross-model benchmarking framework of frontier LLM shopping agents, covering calibration, market concentration, and reasoning effort across ten models and three datasets (\cref{sec:tasks}).

\section{Related Work}
\label{sec:related}

\paragraph{LLM agents for shopping.}
\webshop~\cite{yao2022webshop} introduced a simulated store with 1.18M real products and human instructions and remains the standard testbed for language agents that search, browse and buy; later web-agent benchmarks~\cite{zhou2023webarena,deng2023mind2web} and shopping suites~\cite{jin2024shoppingmmlu} broadened the task mix. These works report end-to-end success, where navigation and decision failures are confounded. \citet{allouah2025agentic} evaluate agents as economic actors and document strong concentration of demand on a few products and sensitivity to listing position, and \citet{elhousni2026aimediation} model the assistant as a mediating layer between retailer and customer and show that it materially changes optimal assortments and revenue. In this paper, we take the complementary approach of investigating the agent's role as a buyer whose decision must be judged, and isolate that decision by fixing the choice set, so every error is attributable to product judgment and conditions are compared on identical inputs. This approach was also advocated for by the recent concurrent work of \citet{candogan2026knowthyagent}, which argues that an agent can complete a purchase while making a poor choice for its user, and evaluates commercial shopping agents on fixed choice sets drawn from a retail catalog against customers with known preferences: \citeauthor{candogan2026knowthyagent} finds that decision quality degrades sharply as choice sets grow and preferences are only partially communicated.

\paragraph{Confidence and calibration of LLMs.}
Classic calibration work in vision~\cite{guo2017calibration,naeini2015ece} measured the gap between a classifier's probability and its accuracy. For LLMs, confidence is usually elicited verbally~\cite{lin2022teaching,tian2023justask,xiong2024canllms} or read from token probabilities~\cite{kadavath2022know}; verbalized confidence is typically overconfident, clustered at a few round values, and only moderately discriminative, as in people~\cite{moore2008overconfidence,tversky1974judgment}. In this paper, we follow this literature in evaluating verbalized confidence with ECE, Brier score and AUROC (defined in \cref{sec:metrics}), but ask how it responds to a reasoning-effort control that was not yet widely adopted (or did not yet exist) when most of the just-mentioned literature was written.

\paragraph{Test-time compute and reasoning effort.}
Chain-of-thought prompting~\cite{wei2022cot} and reinforcement-trained reasoning models~\cite{openai2024o1,deepseek2025r1} established that spending more tokens at inference time can raise accuracy, and scaling studies~\cite{snell2024scaling,muennighoff2025s1} characterized the trade-off. That benefit is typically concentrated in mathematical and symbolic tasks~\cite{sprague2024cot}, and long reasoning can be wasteful on easy problems~\cite{chen2024overthinking}. In this paper, we measure the effect of reasoning effort on agentic shopping in particular and ask whether it changes not only accuracy but also the model's own assessment of that accuracy.

\paragraph{Shared errors across models.}
Error consistency between classifiers has been studied in vision as a signature of shared inductive biases~\cite{geirhos2020error,mania2019similarity}, and recent work warns that increasingly similar LLMs undermine oversight schemes that rely on inter-model disagreement~\cite{goel2025greatmodels}. In this paper, we quantify error sharing in an agentic shopping setting against an explicit independence baseline and show that it is large and insensitive to the thinking effort.

\section{Data and Methods}
\label{sec:method}

\subsection{Three datasets (2 multimodal and 1 text)}
\label{sec:tasks}

All experiments in this study have the same form: a shopping query, a fixed and shuffled list of ten candidate products, and exactly one candidate that is correct. Presenting the full list removes search and navigation from the experiment, so every model at every effort level decides on identical evidence, and it allows a paired analysis in which each model is compared with itself, task by task, across effort levels. The three dataset sources differ in where the query and products come from and in how ``correct'' is defined. We describe each dataset below and provide a summary in \cref{tab:datasets} in the Supplement.

\paragraph{\webshop.}
\webshop~\cite{yao2022webshop} is a multimodal (images with text) dataset whose instructions name a product type and a set of required attributes, option values and a price ceiling (``\emph{I need a long-lasting antiperspirant deodorant for women, 2.6 ounce, under \$40}''). Its reward function (written as a Python function) scores any product--option choice on a scale of 0 to 1 on whether the chosen product's attribute, option and price match what was stated in the query (see Sec. 3 of ~\cite{yao2022webshop}). A choice is \texttt{correct} if and only if it earns the maximum reward of~1. On top of the \webshop environment, we build the following tasks for our experiments: For each given query from the original \webshop test set, we take a target product (that has reward 1 given the query), and we identify nine other products that (a)~are the same kind of product or sharing the target's most specific category, (b)~cannot reach a perfect reward under any option choice and (c)~fail at least one requirement under a lenient paraphrase check. Each product carries its title, a seeded price, bullet features, description, selectable options and main image. Instructions with fewer than nine such {\it distractors} (which we define as products that fulfill the above three requirements) are skipped, leaving 496 tasks (examples in \cref{sec:supp_examples}, \cref{tab:supp_examples} in the Supplement). 

\paragraph{Amazon \sqid.}
\sqid~\cite{reddy2022esci,alghossein2024sqid} combines real Amazon search queries with human judgment-based answers. It is constructed from an earlier Amazon Shopping Queries (ASQ) dataset~\cite{reddy2022esci}, which contains a set of queries and a corresponding set of products for each query. ASQ pairs each query with a list of products that annotators labeled \emph{Exact} (satisfies every specification in the query), \emph{Substitute} (fails some specification but could serve the same purpose), \emph{Complement} (would be used together with an Exact product) or \emph{Irrelevant}. The authors of the original ASQ ~\cite{reddy2022esci} call this ESCI labeling, for Exact, Substitute, Complement, Irrelevant. \sqid adds a product image for most of these products, making it a multimodal dataset. Queries are short and typed the way shoppers type them (``\emph{running sunglasses for men}'', ``\emph{makita inflator 18v}'', ``\emph{kids sleep gummies without melatonin}''; median four words, misspellings kept), so the requirements must be inferred from the query. A choice is \emph{correct} if and only if it is the \texttt{Exact} product, so correctness here is a human verdict from the annotations in ASQ. For the evaluations in this paper, we start from a portion of the \sqid test split, keep products with a product-specific image, and call a query \textit{eligible} if it has at least one \texttt{Exact} and at least nine \texttt{Substitute} or \texttt{Complement} products (3,541 queries). Visiting eligible queries in a seeded random order, we build 500 tasks, each with (a)~one \texttt{Exact} product as the target, chosen at random when the query has several (median five), and (b)~nine distractors from the same query's judged list. More details about the dataset and our evaluation process are in \cref{sec:supp_sq} of the Supplement. Each product carries its title, bullet features, description and image, so the task is pure product identification. Because the ground truth is a subjective human annotation rather than a strict check against objective requirements, some \texttt{Exact} labels are debatable. We treat this label noise as a property of the benchmark: it lowers success for every model and leaves comparisons between models and effort settings intact.

\paragraph{Wayfair \wands.}
\wands~\cite{chen2022wands}, the Wayfair Annotation Dataset, provides 480 real queries from a furniture and home-goods retailer, 42,994 products and 233k human labels that grade each query--product pair as \texttt{Exact}, \texttt{Partial} or \texttt{Irrelevant} under written guidelines. It changes the retailer, the product domain and the modality at once: products come without images, prices or options, and the published text is lowercased and tokenized (``\emph{oletto pull down single handle kitchen faucet}''), so models decide from text alone. Queries are again short search strings (``\emph{full bed with trundle}'', ``\emph{7qt slow cooker}'', ``\emph{3 1/2 inch drawer pull}''; median three words) spanning 159 product classes from accent chairs to bathroom faucets. The correct choice is the Exact product. \wands judges far more products per query than ESCI (median 247, of which about 24 are \texttt{Exact} and 141 \texttt{Partial}), so we can be stricter about distractors: a query is eligible if it has at least one \texttt{Exact} and at least nine \texttt{Partial} products, which 350 of the 480 queries satisfy; 330 form the test set and 20 are held out as a pilot. From the WANDS dataset, we build the following tasks for our experiments: for each of the previous queries, we take (a)~one \texttt{Exact} product at random as the target and (b)~nine \texttt{Partial} products from the same query, i.e.\ products that annotators judged relevant to the query but not a full match, with \texttt{Irrelevant} products never used and (c)~\texttt{Partial} products that share a name with an \texttt{Exact} product dropped as duplicate listings (affecting 47 tasks). Models see the product name as the title, every catalog attribute as a ``\emph{feature: value}'' bullet (median 54 attributes per product with values like ``upholstery material'' and ``overall width'') and the description truncated to 1,500 characters. Product class, category and ratings are not shown. As with \sqid, the ground truth is a human verdict and a few \texttt{Exact} labels are hard to defend (the query ``\emph{almost heaven sauna}'' has as its \texttt{Exact} product a sauna bucket-and-ladle set); such cases again depress absolute success equally for every model and do not affect comparisons.

\subsection{Models and effort manipulation}
\label{sec:models}
We evaluate ten frontier models from six providers: GPT-6 Astra~\cite{openai2026gpt6astra}, GPT-6.1 Sol~\cite{openai2026gpt61sol} and GPT-6 Luna~\cite{openai2026gpt6luna} (OpenAI); Claude Fable 5.1~\cite{anthropic2026fable51} and Claude Opus 5.5~\cite{anthropic2026opus55} (Anthropic); Gemini 4 Argon~\cite{google2026gemini4argon} and Gemini 3.1 Pro~\cite{google2026gemini31pro} (Google); DeepSeek V4.1 Flash~\cite{deepseek2026v41flash}\footnote{We use the DeepSeek ``Flash'' series rather than the higher-visibility ``Pro'' series because the ``Pro'' models are not multimodal.}; Qwen 3.8 Max~\cite{qwen2026qwen38max} (Alibaba); and Kimi K3~\cite{moonshot2026kimik3} (Moonshot). All are queried through the OpenRouter API~\cite{openrouter2026} with identical prompts and a shared output budget of 16,000 tokens for reasoning plus answer. Reasoning effort is set with the OpenRouter's \texttt{reasoning.effort} parameter at \texttt{low}, \texttt{medium}, or \texttt{high}\footnote{We checked that the \texttt{reasoning.effort} parameter is not ignored by the benchmarked models' OpenRouter APIs. Additionally, we note DeepSeek V4.1 and Kimi K3 which allow only \texttt{low} and \texttt{high} reasoning.}.  Sampling settings are the provider defaults.

\paragraph{Prompt and response.}
The system prompt casts the model as a shopping assistant that must buy the single product, with the right options, that best satisfies the customer. The user message gives the query, the ten products (title, price, features, description, options and image where available) and asks for a JSON object with the chosen product number, one value per option type and ``\emph{how confident you are that your chosen product and options satisfy every requirement in the customer instruction, as an integer from 0 (not at all confident) to 10 (completely confident)}''. Of 37,128 responses across the three datasets, 37,099 (99.92\%) contained a parseable choice and confidence; the rest are dropped. The full prompt is in \cref{sec:supp_prompt} of the Supplement.

\subsection{Metrics and statistical analysis}
\label{sec:metrics}

\paragraph{Success rate, confidence and overconfidence.}
We refer to the proportion of correctly-answered queries among all queries answered by a certain LLM at a certain effort level as the \emph{success rate} for that effort level. We will report success rates by model and effort level and 95\% confidence intervals on their paired changes across effort levels. \emph{Confidence} is the integer from 0 to 10 that the model attaches to its answer; we will report its mean per model and effort level with 95\% confidence intervals on its paired changes across effort levels. In figures and tables, we rescale the confidence to 0--100 scaling so that it can be compared to the success rate. Lastly, we let \emph{overconfidence} be the difference between the rescaled confidence and the success rate, which we will also report by model and effort level.

\paragraph{Calibration metrics.}
Since the confidence $c$ is between 1 and 10, we can rescale using $c/10$ and treat it as a probability. The \emph{expected calibration error} (ECE)~\cite{naeini2015ece,guo2017calibration} is the average absolute gap between stated confidence and realized success rate across the eleven confidence levels, weighted by how often each level is used; the \emph{Brier score}~\cite{brier1950} is the mean squared gap between $c/10$ and the 0/1 outcome, which penalizes both miscalibration and inaccuracy; and the \emph{AUROC}~\cite{hanley1982auroc} is the probability that a randomly chosen correct answer carries higher confidence than a randomly chosen wrong one, which measures only whether confidence \emph{ranks} right above wrong, regardless of its level (\cref{sec:calibration}, \cref{tab:calibration}). A model can have a poor ECE and a good AUROC if its confidence is informative but uniformly too high.

\paragraph{Effect of effort.}
For every model and every effort-level pair (low$\rightarrow$medium, medium$\rightarrow$high, low$\rightarrow$high), we take the tasks answered under both settings, compute the within-task difference in success and confidence, and test whether mean difference differs from zero with a paired $t$-test, reporting the mean with a 95\% confidence interval (\cref{sec:effort}). 


\paragraph{Shared errors.}
For each effort level, we take the tasks answered by all $n$ models at that level ($n=10$ at low and high; 8 at medium) and count, for each task, how many of the $n$ models chose wrong (\cref{fig:concentration}), which we denote by $k$. If the models' errors were independent, this count would follow a binomial distribution, $\mathrm{Binomial}(n,\bar p)$, with the mean error rate $\bar p$, tightly concentrated around $n\bar p$. We compare the observed distribution of the count to this independence baseline in three ways: (a) the number of tasks that all models fail or all solve, observed versus expected; (b) the \emph{variance ratio}, the variance of the observed count divided by the variance under independence, which is 1 if errors are independent and grows as errors become shared; and (c) the \emph{pairwise error correlation} $\rho=\bigl(\mathrm{Var}(k)/(\bar p(1-\bar p))-n\bigr)/\bigl(n(n-1)\bigr)$ (where the $\mathrm{Var}$ is taken across tasks) the correlation between any two models' error indicators that would produce the observed variance. We also report the Kullback--Leibler (KL) divergence of the observed distribution from the binomial baseline. Finally, on tasks where at least two models erred, we ask what share of the wrong choices fall on the single most-chosen wrong product, against a Monte-Carlo baseline in which each wrong model picks uniformly among the nine distractors (\cref{sec:concentration}).

\section{Results}
\label{sec:results}

Due to page limit constraints, some figures in this section show only results from the \webshop dataset; the corresponding figures for \sqid and \wands are in \cref{sec:supp_sq,sec:supp_wa} of the Supplement and are referenced where the text discusses them. The tables cover all three datasets.

\begin{figure*}[t]
\centering
\includegraphics[width=0.8\textwidth]{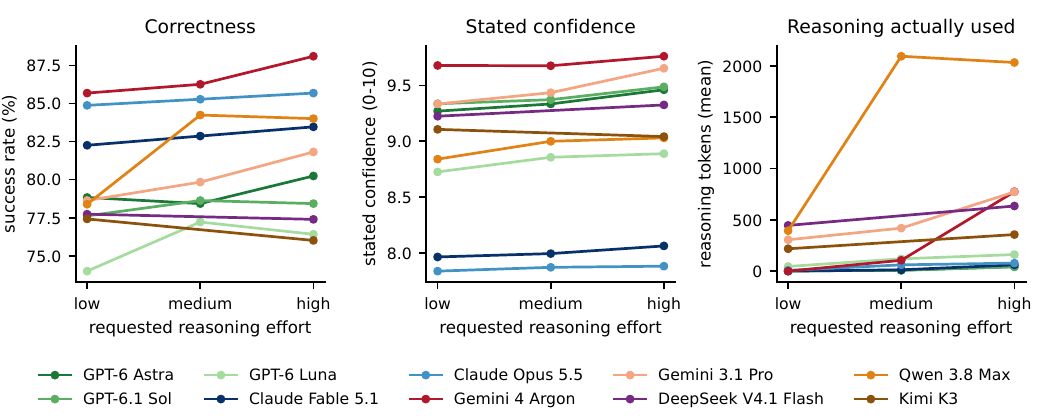}
\caption{\textbf{\webshop at a glance.} Per model, the requested reasoning effort is varied from low to high. The graphs display success rate (left), mean stated confidence (middle) and mean reasoning tokens actually used (right) for each reasoning level. All 28 \webshop conditions lie between 74.0\% and 88.1\% success. We see that Qwen 3.8 Max goes from 394 to 2,033 reasoning tokens per task between low and high effort (488 to 3,926 on \sqid), DeepSeek V4.1 Flash from 445 to 634, Gemini 3.1 Pro from 305 to 772 and Gemini 4 Argon from 0 to 774, whereas GPT-6 Astra and GPT-6.1 Sol stay below 100 reasoning tokens per task even at high effort. Similar graphs for \sqid and \wands can be found in \cref{fig:supp_grid}. \textbf{Takeaway: } Success rate does not rise significantly from low to high reasoning for most models, while the model output's stated confidence does.}
\label{fig:grid}
\end{figure*}

\subsection{Overview}
\label{sec:overview}

\Cref{fig:grid} and \cref{fig:supp_grid} show every model's success, confidence and reasoning-token usage at the three effort levels on \webshop. We see that success lines are nearly flat and cross each other, confidence lines are nearly all rising, and token usage can differ by two orders of magnitude between providers. Within a dataset, models are spread over 4--11 points, and the ordering is only partly stable: success ranks agree closely between \webshop and \sqid, but not between either of \webshop-or-\sqid and \wands. Perhaps this is because \wands is text-only while \webshop and \sqid are multimodal with both images and text.

The right panel of \cref{fig:grid} shows that reasoning effort affects the actual number of tokens a model uses very unevenly. On \webshop, models like Qwen 3.8 Max and the Gemini models vary more drastically in number of reasoning tokens for each effort level. In contrast, GPT-6 Astra, GPT-6.1 Sol and both Claude models only slightly increase their token usage with higher reasoning efforts. The same pattern holds on the other two datasets.

The middle panel of \cref{fig:grid} shows a gap in stated confidence that recurs throughout our results. On \webshop, the two Claude models report a mean confidence of about 8 out of 10, while the other eight models report between 8.7 and 9.7. Measured against their success rates, those eight models are overconfident by 7 to 16 percentage points, whereas the two Claude models are slightly \emph{under}confident ($-2.8$ and $-6.7$ points for Claude Fable 5.1 and Claude Opus 5.5). This gap will reappear later in the calibration metrics (\cref{sec:calibration}). 

\subsection{Raising effort changes confidence, not accuracy}
\label{sec:effort}

\begin{figure*}[t]
\centering
\includegraphics[width=0.9\textwidth]{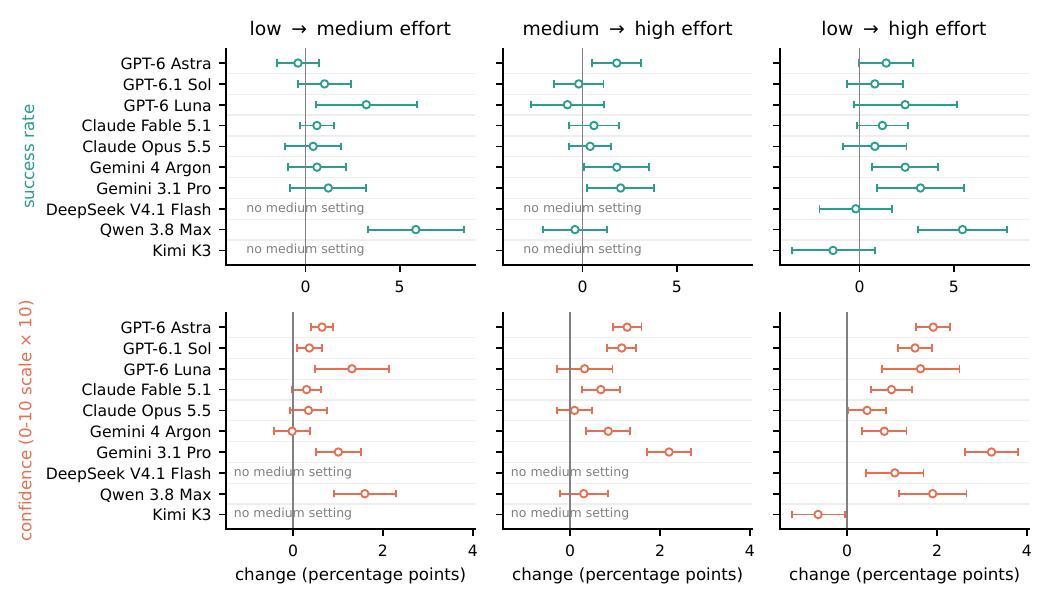}
\caption{\textbf{Paired within-task change on \webshop when requested effort is changed.} The 95\% confidence intervals for each case are added. The columns are low$\rightarrow$medium, medium$\rightarrow$high, and low$\rightarrow$high. The top row shows change in success rate (percentage points), and the bottom row shows change in mean stated confidence (0--10 scale, $\times 10$ so the units match). DeepSeek V4.1 Flash and Kimi K3 have no medium setting. Graphs for \sqid and \wands can be found in \cref{fig:supp_effects_sq,fig:supp_effects_wa}. \textbf{Takeaway:} Most accuracy intervals cover zero (indicating no significant change in accuracy), and confidence intervals mostly sit above it (indicating an increase in confidence).}
\label{fig:effects}
\end{figure*}

\begin{table}[t]
\centering
\caption{\textbf{Paired low$\rightarrow$high change in success rate (percentage points, teal) and mean stated confidence (0--10 scale $\times 10$, coral) on the three datasets.} Shaded bold cells differ significantly from zero in a paired $t$-test at $p<0.05$. A dagger ($^{\dagger}$) marks changes that remain significant ($p<0.05$) after Holm correction~\cite{holm1979} for the 26 contrasts per metric and dataset; an asterisk ($^{*}$) marks changes significant only before correction. Holm correction is strictly more conservative than the uncorrected test (\cref{sec:metrics}), so every $^{\dagger}$ cell would also pass the uncorrected test. The contrasts from low$\rightarrow$medium and medium$\rightarrow$high per dataset can be found in \cref{tab:supp_effects_ws,tab:supp_effects_sq,tab:supp_effects_wa}. \textbf{Takeaway:} Increasing thinking level increases confidence but typically does not improve success rate in a statistically significant way.}
\label{tab:effects}
\footnotesize
\setlength{\tabcolsep}{3pt}
\resizebox{\linewidth}{!}{\begin{tabular}{l>{\columncolor{accbg}}c>{\columncolor{accbg}}c>{\columncolor{accbg}}c!{\vrule width 1.2pt}>{\columncolor{confbg}}c>{\columncolor{confbg}}c>{\columncolor{confbg}}c}
\toprule
& \multicolumn{3}{c!{\vrule width 1.2pt}}{\cellcolor{accbg}\textbf{Success rate (pts)}} & \multicolumn{3}{c}{\cellcolor{confbg}\textbf{Stated confidence (pts)}} \\
Model & WebShop & SQID & WANDS & WebShop & SQID & WANDS \\
\midrule
GPT-6 Astra & +1.4 & +0.8 & +0.9 & \cellcolor{confsig}\textbf{+1.9$^{\dagger}$} & \cellcolor{confsig}\textbf{+2.5$^{\dagger}$} & \cellcolor{confsig}\textbf{+1.7$^{\dagger}$} \\
GPT-6.1 Sol & +0.8 & -0.2 & +0.9 & \cellcolor{confsig}\textbf{+1.5$^{\dagger}$} & \cellcolor{confsig}\textbf{+2.4$^{\dagger}$} & \cellcolor{confsig}\textbf{+1.9$^{\dagger}$} \\
GPT-6 Luna & +2.4 & +2.0 & \cellcolor{accsig}\textbf{+5.2$^{*}$} & \cellcolor{confsig}\textbf{+1.6$^{\dagger}$} & +0.2 & \cellcolor{confsig}\textbf{+1.3$^{*}$} \\
Claude Fable 5.1 & +1.2 & +1.2 & -1.8 & \cellcolor{confsig}\textbf{+1.0$^{\dagger}$} & +0.5 & +0.5 \\
Claude Opus 5.5 & +0.8 & +0.2 & -2.1 & \cellcolor{confsig}\textbf{+0.4$^{*}$} & +0.4 & -0.3 \\
Gemini 4 Argon & \cellcolor{accsig}\textbf{+2.4$^{*}$} & -0.6 & +0.9 & \cellcolor{confsig}\textbf{+0.8$^{\dagger}$} & +0.1 & +0.2 \\
Gemini 3.1 Pro & \cellcolor{accsig}\textbf{+3.2$^{*}$} & -0.4 & -0.6 & \cellcolor{confsig}\textbf{+3.2$^{\dagger}$} & \cellcolor{confsig}\textbf{+1.9$^{\dagger}$} & \cellcolor{confsig}\textbf{+4.5$^{\dagger}$} \\
DeepSeek V4.1 Flash & -0.2 & -2.0 & +0.0 & \cellcolor{confsig}\textbf{+1.1$^{\dagger}$} & \cellcolor{confsig}\textbf{+2.0$^{\dagger}$} & \cellcolor{confsig}\textbf{+2.3$^{\dagger}$} \\
Qwen 3.8 Max & \cellcolor{accsig}\textbf{+5.5$^{\dagger}$} & +1.6 & +0.6 & \cellcolor{confsig}\textbf{+1.9$^{\dagger}$} & \cellcolor{confsig}\textbf{+1.1$^{*}$} & \cellcolor{confsig}\textbf{+2.0$^{\dagger}$} \\
Kimi K3 & -1.4 & -0.4 & +1.5 & \cellcolor{confsig}\textbf{-0.6$^{*}$} & -0.8 & \cellcolor{confsig}\textbf{-0.8$^{*}$} \\
\bottomrule
\end{tabular}
}
\end{table}

\Cref{fig:effects} plots, for every pair of LLM effort levels, the mean within-task change in success (top) and confidence (bottom) with 95\% intervals in \webshop. \cref{tab:effects} gives the low$\rightarrow$high change on all three datasets with significance markers.

\paragraph{Accuracy.}
As we can see from \cref{fig:effects}, across the 30 model--dataset pairs, only three low$\rightarrow$high changes are significant at the nominal level on \webshop: Qwen 3.8 Max, Gemini 3.1 Pro, and Gemini 4 Argon. The same graphs for the other datasets (\cref{fig:supp_effects_sq,fig:supp_effects_wa}) show that on \wands, GPT-6 Luna is the only significant low$\rightarrow$high increase, and no models significantly increase their accuracy on \sqid. Out of these, only Qwen on \webshop survives Holm correction~\cite{holm1979} (see \cref{tab:effects} caption) for the 26 comparisons per dataset. Additionally, GPT-6 Luna survives Holm correction the for the low$\rightarrow$medium increase.

\paragraph{Confidence.}
The bottom row of \cref{fig:effects} looks different. From low to high effort, confidence rises significantly in the low$\rightarrow$high case for nine of ten models on \webshop. From the other datasets (\cref{fig:supp_effects_sq,fig:supp_effects_wa}), it rises significantly on five of ten on \sqid and six of ten on \wands. It only falls significantly for Kimi K3 (on \webshop and \wands) for low to high effort. Counting all contrasts, confidence changes significantly in 20, 11, and 14 (of 26 comparisons) in \webshop, \sqid, and \wands, respectively (17, 9, and 12 after correction) with almost all of these changes being upward. The increases are modest on the 0--10 scale, but are still systematic and are usually not accompanied by accuracy gains. Furthermore, for eight models they start from a level that is already too high. Gemini 3.1 Pro is the clearest case: its confidence rises by a significant amount on all three datasets while its accuracy only rises on \webshop. Its \webshop AUROC actually falls, while the share of answers rated 10/10 rises heavily. Consistent with this, low$\rightarrow$high changes in ECE are within a few points most of the model--dataset pairs. The only significant calibration improvements are Qwen on \webshop and GPT-6 Luna on \wands, both of which had their ECE decrease. In both of these cases, this was only because their accuracy caught up with an already high confidence.


\subsection{Calibration is a model property, not the setting}
\label{sec:calibration}

\begin{table*}[t]
\centering
\caption{\textbf{Calibration by model on the three datasets (averaged over effort levels).} ECE is measured in percentage points, Brier score is measured on a $c/10$ scale, and AUROC is measured on a $0-1$ scale. Lower ECE and Brier and higher AUROC are better, and the best value per column is bold. Per-effort breakdowns are shown in \cref{tab:supp_cal_ws,tab:supp_cal_sq,tab:supp_cal_wa}. On ECE and Brier the Claude models lead by a wide margin everywhere (with the exception of \webshop Brier scores, where Gemini 4 Argon does better): ECE 3.4 and 6.7 points against 8.9--18.2 for the other eight on \webshop, 13.2 and 12.4 against 27.6--34.3 on \sqid, 5.0 and 5.3 against 11.8--19.9 on \wands. \textbf{Takeaway:} The relative calibration rankings are a model property rather than a dataset property: models are ordered nearly identically by ECE on all three datasets (Spearman correlation $\rho=0.79$--$0.92$) despite a large spread in accuracy.}
\label{tab:calibration}
\small
\setlength{\tabcolsep}{5pt}
\begin{tabular}{l>{\columncolor{ecebg}}c>{\columncolor{ecebg}}c>{\columncolor{ecebg}}c!{\vrule width 1.2pt}>{\columncolor{brierbg}}c>{\columncolor{brierbg}}c>{\columncolor{brierbg}}c!{\vrule width 1.2pt}>{\columncolor{aurocbg}}c>{\columncolor{aurocbg}}c>{\columncolor{aurocbg}}c}
\toprule
& \multicolumn{3}{c!{\vrule width 1.2pt}}{\cellcolor{ecebg}\textbf{ECE (pts) $\downarrow$}} & \multicolumn{3}{c!{\vrule width 1.2pt}}{\cellcolor{brierbg}\textbf{Brier $\downarrow$}} & \multicolumn{3}{c}{\cellcolor{aurocbg}\textbf{AUROC $\uparrow$}} \\
Model & WebShop & SQID & WANDS & WebShop & SQID & WANDS & WebShop & SQID & WANDS \\
\midrule
GPT-6 Astra & 17.5 & 33.1 & 19.1 & 0.196 & 0.350 & 0.214 & 0.56 & 0.67 & 0.68 \\
GPT-6.1 Sol & 18.2 & 33.9 & 19.9 & 0.203 & 0.354 & 0.217 & 0.56 & 0.67 & 0.69 \\
GPT-6 Luna & 15.2 & 30.4 & 14.8 & 0.202 & 0.324 & 0.198 & 0.59 & 0.69 & 0.76 \\
Claude Fable 5.1 & \textbf{3.4} & 13.2 & \textbf{5.0} & 0.135 & 0.225 & 0.155 & 0.69 & 0.74 & 0.80 \\
Claude Opus 5.5 & 6.7 & \textbf{12.4} & 5.3 & 0.126 & \textbf{0.225} & \textbf{0.155} & \textbf{0.69} & 0.73 & \textbf{0.81} \\
Gemini 4 Argon & 10.8 & 30.1 & 18.6 & \textbf{0.119} & 0.320 & 0.196 & 0.66 & 0.69 & 0.75 \\
Gemini 3.1 Pro & 16.5 & 32.0 & 16.4 & 0.181 & 0.335 & 0.187 & 0.61 & 0.70 & 0.77 \\
DeepSeek V4.1 Flash & 16.8 & 34.3 & 18.5 & 0.201 & 0.344 & 0.205 & 0.57 & 0.72 & 0.78 \\
Qwen 3.8 Max & 8.9 & 27.6 & 11.8 & 0.149 & 0.290 & 0.179 & 0.63 & \textbf{0.74} & 0.76 \\
Kimi K3 & 15.1 & 32.5 & 15.5 & 0.199 & 0.331 & 0.188 & 0.59 & 0.71 & 0.76 \\
\bottomrule
\end{tabular}

\end{table*}

\Cref{tab:calibration} reports ECE, Brier and AUROC (\cref{sec:metrics}) for every model on every dataset; the per-effort breakdown is in \cref{tab:supp_cal_ws}, \cref{tab:supp_cal_sq}, and \cref{tab:supp_cal_wa} of the Supplement. As we can see, the Claude models seem to be the best calibrated across all metrics. On ECE and Brier the Claude models lead by a wide margin everywhere (with the exception of \webshop Brier scores, where only Gemini 4 Argon does better). In overconfidence terms, the Claude models are $-2.8$ and $-6.7$ points on \webshop, $+0.7$ and $-1.5$ on \wands, and $+12.4$ and $+10.8$ on \sqid, where every model's accuracy falls to around 50\% while stated confidence falls much less; the other eight are $+7$ to $+16$, $+10$ to $+20$ and $+27$ to $+33$ points overconfident on the three datasets. The rankings of models by overconfidence and by ECE are nearly identical across datasets whose accuracy levels differ by forty points.

On AUROC, which ignores the level of confidence, the gap narrows. The Claude models are first on \webshop and \wands, and on \sqid Qwen 3.8 Max matches them while most models cluster at a slightly lower level. The overconfident models therefore do carry some information in their confidence; they report it on a scale shifted upward by 10--30 points. AUROC is higher on the two human-labeled datasets than on \webshop for every model.

Per-effort tables (\cref{tab:supp_cal_ws,tab:supp_cal_sq,tab:supp_cal_wa}) in the Supplement that show that changing reasoning effort does not change any of these conclusions. Reading across the \texttt{low}, \texttt{medium}, and \texttt{high} effort columns in the tables, ECE moves by less than two points for most model--dataset pairs and by less than five for all but the cases already noted (Qwen and GPT-6 Luna improving with their accuracy gains), and AUROC barely moves everywhere except Gemini 3.1 Pro on \webshop.

\subsection{Models fail together}
\label{sec:concentration}

\begin{figure*}[t]
\centering
\includegraphics[width=0.8\textwidth]{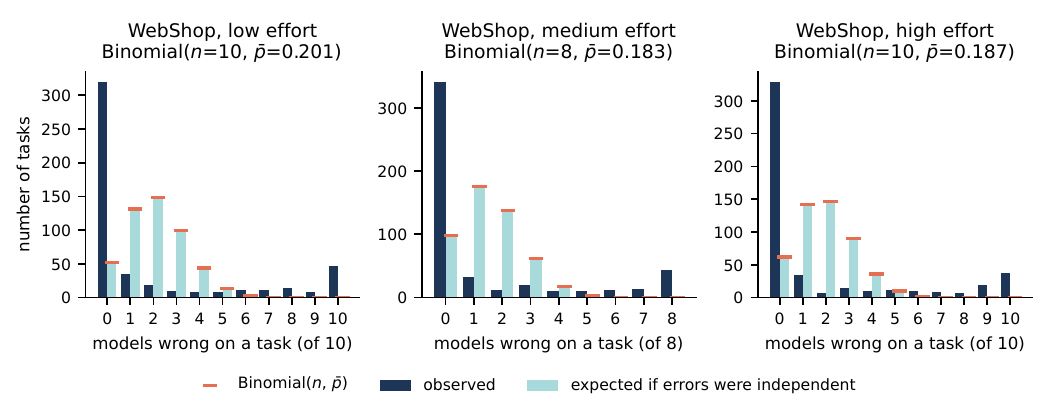}
\caption{\textbf{\webshop: how many models are wrong on the same task}. The graphs show at this and what independent errors would produce at each effort level. The dark bars are the observed number of tasks on which exactly $k$ models chose wrong, and the light bars are the expected count if the models' errors were independent, $\mathrm{Binomial}(n,\bar p)$. The markers show the same binomial baseline. The medium panel covers the eight models with a medium setting. On \webshop at low effort, this distribution is bimodal: 319 of 493 tasks are solved by all ten models (independence predicts 50) and 47 are failed by all ten (independence predicts 0.0001). Similar graphs for \sqid and \wands can be found at \cref{fig:supp_conc_sq,fig:supp_conc_wa}. \textbf{Takeaway:} At every effort level, the bimodal distribution of the errors indicates that models tend to fail together on the same tasks.}
\label{fig:concentration}
\end{figure*}

\begin{table}[t]
\centering
\caption{\textbf{Shared errors on the three datasets, by effort level.} For each effort level, there are $n$ models, with mean error rate $\bar p$. We show the observed vs.\ independence-expected number of tasks on which models all-were-wrong or all-were-right, the variance ratio (observed variance of the number of wrong models over its variance under independence), the implied pairwise error correlation $\rho$, and the KL divergence of the observed distribution from $\mathrm{Binomial}(n,\bar p)$. The pairwise error correlation implied by the variance is 0.70--0.71 on \webshop, 0.79--0.81 on \sqid and 0.77--0.81 on \wands. \textbf{Takeaway:} The variance of the number of wrong models is far more than in an independent-models baseline.}
\label{tab:concentration}
\footnotesize
\setlength{\tabcolsep}{3pt}
\resizebox{\linewidth}{!}{\begin{tabular}{llcccccc}
\toprule
Dataset & Effort & $n$, $\bar p$ & All wrong & All right & Var ratio & $\rho$ & KL div. \\
& & & obs / indep & obs / indep & & & \\
\midrule
\multirow{3}{*}{WebShop} & low & 10, 0.201 & 47 / 0.0 & 319 / 51.8 & 7.4 & 0.71 & 2.65 \\
 & medium & 8, 0.183 & 43 / 0.0 & 340 / 97.0 & 5.9 & 0.70 & 1.90 \\
 & high & 10, 0.187 & 37 / 0.0 & 328 / 61.1 & 7.4 & 0.71 & 2.54 \\
\midrule
\multirow{3}{*}{SQID} & low & 10, 0.528 & 185 / 0.8 & 171 / 0.3 & 8.1 & 0.79 & 4.12 \\
 & medium & 8, 0.519 & 194 / 2.6 & 185 / 1.4 & 6.6 & 0.80 & 3.21 \\
 & high & 10, 0.525 & 189 / 0.8 & 173 / 0.3 & 8.3 & 0.81 & 4.24 \\
\midrule
\multirow{3}{*}{WANDS} & low & 10, 0.285 & 55 / 0.0 & 184 / 11.4 & 7.9 & 0.77 & 3.48 \\
 & medium & 8, 0.282 & 62 / 0.0 & 205 / 23.3 & 6.7 & 0.81 & 2.88 \\
 & high & 10, 0.280 & 55 / 0.0 & 204 / 12.3 & 8.3 & 0.81 & 3.76 \\
\bottomrule
\end{tabular}
}
\end{table}

If the ten models made their mistakes independently, the number of models wrong on a given task would look like a binomial distribution. However, \Cref{fig:concentration} shows that the observed distribution looks nothing like this. It is instead bimodal, and we see that the observed counts pile up at ``none wrong'' and ``all wrong'' instead of clustering around the average. Tasks are more often either correctly solved by all ten models or failed by all ten than somewhere in between. The same shape appears at medium and high effort and on the other two datasets (\cref{fig:supp_conc_sq,fig:supp_conc_wa}).

\Cref{tab:concentration} quantifies this. The variance of the number of wrong models is 6--8 times what independence would produce, and the pairwise error correlation implied by that variance is high on all three datasets. None of these quantities moves with effort\footnote{The variance ratio and KL divergence are lower at medium only because DeepSeek V4.1 and Kimi K3 do not have medium reasoning.} and the error correlation $\rho$, which is normalized for panel size, changes only minimally.

Moreover, models do not merely fail on the same tasks, they fail in the same way. On tasks where at least two models chose wrong at low effort, the share of wrong choices landing on the single most popular wrong product is over 80\% on all datasets, against 32--36\% if wrong models picked uniformly among the nine distractors. In the \webshop, the shared wrong product is typically a plausible near miss that satisfies the product type and a headline attribute but fails a secondary constraint such as size, scent or price. On \sqid, it is most often a \texttt{Substitute} whose title is nearly identical to the \texttt{Exact} product's. Practically, this suggests that agreement among frontier models is weak evidence that a shopping decision is correct, and an ensemble or ``second opinion'' recovers much less than independent errors would suggest.

\section{Discussion}
\label{sec:discussion}

\paragraph{Effort buys confidence, not correctness.}
In this paper, we show that increasing reasoning effort increases confidence but not correctness holds across three agentic shopping benchmarks: WebShop, SQID, and WANDS. One explanation is that product choice is bounded by reading rather than reasoning: the information needed to answer is in the listing, and a model that misses a secondary constraint or is fooled by a near-identical substitute at low effort does not recover it by deliberating longer. A second explanation is that a long reasoning trace supplies the model with the \emph{experience} of having checked, which it then reports as confidence. The two cases in which accuracy did respond, Qwen 3.8 Max on \webshop (\cref{fig:effects}) and GPT-6 Luna on \wands (\cref{fig:supp_effects_wa} in the Supplement) both delivered the whole gain at the low$\rightarrow$medium step, which suggests that there are tasks sufficiently difficult where, for certain models, the low setting is genuinely insufficient, an increase in effort does help, but the improvement plateaus immediately afterwards with enough thinking at medium.

\paragraph{Calibration is a model property.}
The ranking of models by overconfidence barely changes between different datasets between which accuracy ranges often differ by almost forty percentage points (\cref{fig:supp_compare} in the Supplement). 
Users and platform builders should therefore treat calibration as a selection criterion rather than a tuning target.

\paragraph{Shared errors limit oversight.}
Pairwise error correlations of 0.7--0.8 mean that two frontier agents disagreeing is rare and two agreeing is nearly uninformative (\cref{fig:concentration,tab:concentration}). For the consumer, the agent's confidence and a second agent's opinion are both weak checks; for the market, correlated errors mean that when agents are systematically drawn to a particular near-miss listing, that listing captures demand across all agents at once, which is the concentration \citet{allouah2025agentic} observe from the seller's side and the mediation effect \citet{elhousni2026aimediation} model. Reasoning effort does not break this correlation: the models are as aligned in their mistakes at high effort as at low.

\paragraph{Implications for evaluation.}
Effort settings are often believed to improve the accuracy of decision tasks. However, in agentic shopping, increasing thinking effort doe little for accuracy but significantly increases the models' reported confidence. Thus, evaluations of agentic systems should report calibration and effort-conditioned results alongside accuracy, use paired designs with multiple-comparison control (8 of our 11 nominally significant accuracy effects disappear under correction; see \cref{tab:supp_summary} in the Supplement), and state how many reasoning tokens a setting actually produced, since for two providers ``high effort'' means fewer than 100.

\section{Limitations}
\label{sec:limitations}

Our tasks isolate the decision by handing the model a fixed ten-product list; they do not test search, pagination or multi-step navigation, where effort might matter more. Ground truth is imperfect: about 15\% of \webshop distractors are arguably acceptable, and the \sqid and \wands labels carry the known noise of crowd relevance judgements, which depresses absolute accuracy for every model and may inflate the apparent error correlation, since a mislabeled task counts every model as wrong. Comparisons across models and effort levels within a dataset share the same labels and are affected much less. Confidence was elicited once, as an integer 0--10 in the same JSON as the answer; other elicitation protocols~\cite{xiong2024canllms,tian2023justask} might yield different levels, though the stability of rankings across three datasets suggests the model differences are robust. Each condition was run once with provider-default sampling, so run-to-run variance is absorbed into the task-level noise of the paired tests rather than estimated separately. Finally, the effort control is a provider-defined abstraction whose implementation varies, as the token counts show; our results describe the control as users experience it, not a fixed amount of computation.

\section{Conclusion}
\label{sec:conclusion}

We built evaluations for agentic shopping from three benchmark datasets. We then asked ten frontier LLMs to choose and rate their confidence at \texttt{low}, \texttt{medium}, and \texttt{high} reasoning levels. On every dataset, effort had little effect on accuracy while it significantly raised stated confidence for several models; calibration was a stable attribute of the model, with two models near-calibrated or slightly overconfident and the rest substantially overconfident at every setting; and models failed together far more often than chance, usually on the same wrong product. Thus, for agentic shopping, choosing a calibrated model matters more than turning up its effort, and neither its confidence nor its agreement with other agents is a reliable guard against error. Future work should test whether the pattern persists when the agent also searches, whether training for calibration transfers to this setting, and whether the shared errors we observe stem from shared training data, shared listing cues, or both.

\vfill
\pagebreak
\section*{LLM Usage Statement}
\label{sec:llm}

Large language models were used in this work in two distinct roles. First, as the \emph{objects of study}: the ten commercial models listed in \cref{sec:models} were queried through the OpenRouter API to produce the product choices, confidence ratings and reasoning-token counts analyzed in the paper; their outputs are data, not contributions to the manuscript. Second, as a \emph{research and writing aid}: an LLM-based coding assistant (Anthropic's Claude, accessed through the Cursor editor) was used to help write and debug the Python code for task construction, scoring, statistical analysis and figure and table generation, and to draft and edit the text of this paper, including the related-work discussion, from the authors' experimental design, results and notes. All experimental decisions, the analysis plan, the interpretation of results and the final text were made and verified by the authors, who reviewed every number against the raw data and take full responsibility for the content. No generative AI tool was used to create or alter images or to fabricate data or citations.

{
    \small
    \bibliographystyle{ieeenat_fullname}
    \bibliography{references}
}


\clearpage
\setcounter{page}{1}

\setcounter{section}{0}
\renewcommand{\thesection}{\Alph{section}}
\renewcommand{\theHsection}{supp.\Alph{section}}
\setcounter{figure}{0}
\renewcommand{\thefigure}{S\arabic{figure}}
\setcounter{table}{0}
\renewcommand{\thetable}{S\arabic{table}}

\setcounter{dbltopnumber}{4}
\renewcommand{\dblfloatpagefraction}{0.75}
\raggedbottom

\twocolumn[{%
\renewcommand\twocolumn[1][]{#1}%
\maketitlesupplementary
\begin{center}
\captionsetup{type=table}
\captionof{table}{\textbf{The three datasets used in this experiment.} Each experiment for each dataset has ten candidates and exactly one correct product. Incorrect products are always near misses drawn from the same query, never random products. Construction details for each benchmark are given in \cref{sec:tasks}.}
\label{tab:datasets}
\small
\setlength{\tabcolsep}{8pt}
\begin{tabular}{lccc}
\toprule
 & \webshop~\cite{yao2022webshop} & \sqid~\cite{reddy2022esci,alghossein2024sqid} & \wands~\cite{chen2022wands} \\
\midrule
Source catalog & Amazon (WebShop) & Amazon (ESCI) & Wayfair \\
Query type & attribute instruction & search query & search query \\
Ground truth & simulator reward & human Exact / Substitute / Complement & human Exact / Partial \\
Distractors & same kind of product, reward $<1$ & Substitutes, then Complements & Partial matches \\
Images shown & yes & yes & no \\
Price / options shown & yes / yes & no / no & no / no \\
Tasks & 496 & 500 & 330 \\
Valid decisions & 13,875 & 13,990 & 9,234 \\
\bottomrule
\end{tabular}
\end{center}
\vspace{1em}
}]

\section{Summary across datasets}
\label{sec:supp_summary}

\Cref{tab:datasets} gives an overview of the three datasets we use in this experiment.

\begin{figure*}[tp]
\centering
\includegraphics[width=\textwidth]{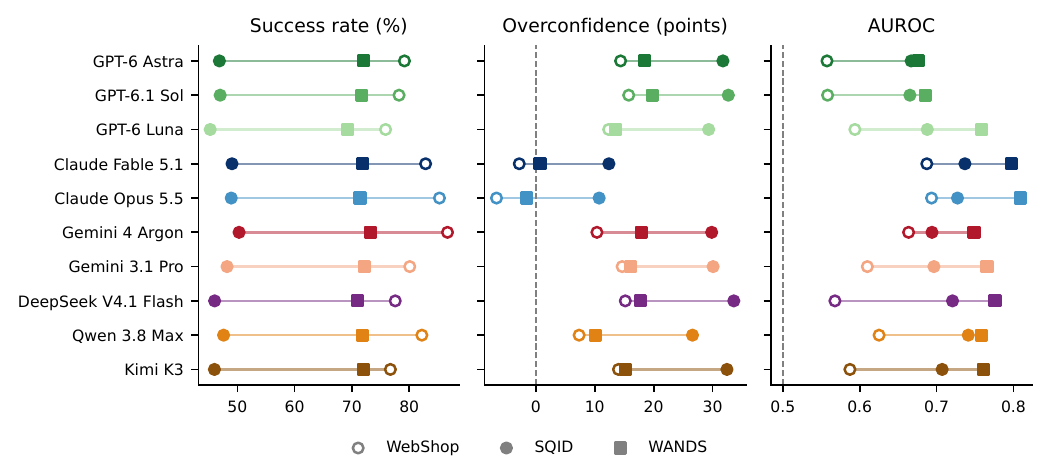}
\caption{\textbf{Per-model success, overconfidence and AUROC on the three datasets} (averaged over effort; open circle \webshop, filled circle \sqid, square \wands). Success differs mainly by dataset; overconfidence separates the models into the same groups on every dataset.}
\label{fig:supp_compare}
\end{figure*}

\Cref{tab:supp_summary} collects the headline statistics of the main paper for the three datasets in one place, and \cref{fig:supp_compare} plots per-model success, overconfidence and AUROC on the three datasets side by side.

\begin{table}[ht]
\centering
\caption{\textbf{Headline results on the three datasets.} Rows are defined in the main paper; Claude values are Fable 5.1 / Opus 5.5.}
\label{tab:supp_summary}
\footnotesize
\setlength{\tabcolsep}{3pt}
\resizebox{\linewidth}{!}{\begin{tabular}{lccc}
\toprule
& \webshop & \sqid & \wands \\
\midrule
Tasks / valid decisions & 496 / 13,875 & 500 / 13,990 & 330 / 9,234 \\
Ground truth & simulator & human labels & human labels \\
Images shown & yes & yes & no \\
\midrule
Success rate, all conditions (\%) & 74.0--88.1 & 43.7--50.6 & 65.8--73.7 \\
Largest $|$low$\rightarrow$high$|$ success change (pts) & 5.5 & 2.0 & 5.2 \\
Success contrasts sig.\ (raw / Holm), of 26 & 8 / 2 & 1 / 0 & 2 / 1 \\
Confidence contrasts sig.\ (raw / Holm), of 26 & 20 / 17 & 11 / 9 & 14 / 12 \\
\quad of which upward & 19 & 11 & 12 \\
\midrule
Overconfidence, Claude Fable / Opus (pts) & -2.8 / -6.7 & +12.4 / +10.8 & +0.7 / -1.5 \\
Overconfidence, other eight models (pts) & +7.4--+15.8 & +26.6--+33.6 & +10.1--+19.8 \\
ECE, Claude Fable / Opus (pts) & 3.4 / 6.7 & 13.2 / 12.4 & 5.0 / 5.3 \\
ECE, other eight models (pts) & 8.9--18.2 & 27.6--34.3 & 11.8--19.9 \\
AUROC, best model & Claude Opus 5.5 & Qwen 3.8 Max & Claude Opus 5.5 \\
Confidence drop when wrong (0--10) & 0.33--0.81 & 1.37--1.85 & 1.26--2.10 \\
\midrule
Pairwise error correlation $\rho$ (low / high) & 0.71 / 0.71 & 0.79 / 0.81 & 0.77 / 0.81 \\
All models wrong, obs.\ / indep.\ (low) & 47 / 0.0 & 185 / 0.8 & 55 / 0.0 \\
Share on modal wrong product (random) & 87\% (36\%) & 81\% (32\%) & 82\% (33\%) \\
\bottomrule
\end{tabular}
}
\end{table}

\section{\webshop: additional material}
\label{sec:supp_ws}

\textbf{Distractor check.} The authors manually checked 100 random distractors and found 85 that clearly violate their instruction and 15 that are arguably acceptable (95\% CI 9--23\%), mostly paraphrases such as ``24-hour protection'' for ``long lasting'' which the lenient paraphrase check did not catch; such cases lower absolute success for every model.

\begin{figure*}[tp]
\centering
\begin{minipage}[t]{0.48\textwidth}
\vspace{0pt}
\centering
\includegraphics[width=\linewidth]{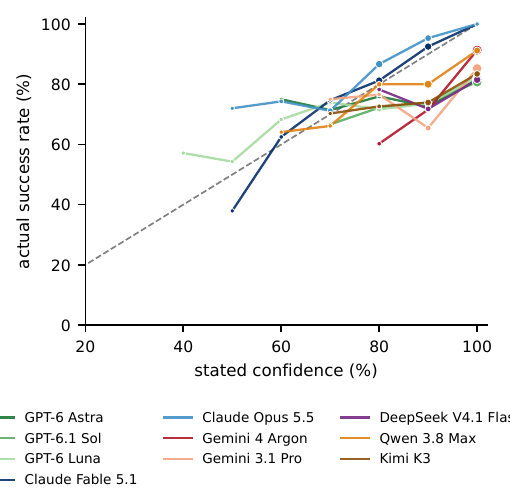}
\caption{\textbf{\webshop reliability diagram.} Realized success rate against stated confidence, per model, averaged over effort levels. The marker size is proportional to the number of answers at that level. The dashed line is perfect calibration.}
\label{fig:supp_rel_ws}
\end{minipage}\hfill
\begin{minipage}[t]{0.48\textwidth}
\vspace{0pt}
\centering
\includegraphics[width=\linewidth]{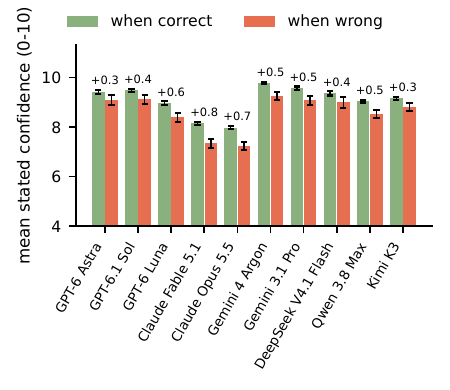}
\caption{\textbf{\webshop: mean stated confidence on correct versus wrong answers}. The graph shows this per model, averaged over effort levels, and with 95\% confidence intervals. The number above each pair is the drop in mean confidence when the model is correct versus wrong. Similar graphs for \sqid and \wands can be found in \cref{fig:supp_rw_sqwa}. We see that every model is more confident when right, but only the Claude models drop by more than 0.6 and only they fall below 8/10 when wrong. They drop by 0.81 and 0.73 against 0.33--0.58 for the rest. Outside the Claude models, mean confidence on \emph{wrong} answers is at least 8.4/10 on \webshop, while for the Claude models, their wrong \webshop answers average 7.2--7.3. \textbf{Takeaway:} Although every model is more confident when right, the Claude models drop the most and have the lowest confidence when wrong.}
\label{fig:rightwrong}
\end{minipage}
\end{figure*}

\Cref{tab:supp_glance_ws} lists every \webshop condition with the Wilson 95\% interval on success, mean confidence, overconfidence, the share of answers rated 10/10, ECE and mean reasoning tokens. \Cref{tab:supp_effects_ws} gives all three paired effort contrasts and \cref{tab:supp_cal_ws} the calibration metrics at each effort level. \Cref{fig:supp_rel_ws} is the reliability diagram: for each model, the realized success rate at each stated confidence level (levels with fewer than 20 answers omitted). Curves below the diagonal indicate overconfidence; the two Claude models are near the diagonal and are generally slightly underconfident (except Fable 5.1 at $50\%$ confidence bin), the others lie 10--20 points below it at their most-used confidence levels. \Cref{fig:rightwrong} is discussed in \cref{sec:rightwrong}.

\begin{table*}[tp]
\centering
\caption{\textbf{\webshop results per condition.} Success in percent with Wilson 95\% CI; mean stated confidence; overconfidence in points; share of answers rated 10/10; ECE in points; mean reasoning tokens.}
\label{tab:supp_glance_ws}
\small
\begin{tabular}{llcccccc}
\toprule
Model & Effort & Success (95\% CI) & Conf. & Overconf. & 10/10 & ECE & Tokens \\
\midrule
\multirow{3}{*}{GPT-6 Astra} & low & 78.8 (75--82) & 9.27 & +13.9 & 71\% & 18.0 & 3 \\
 & medium & 78.4 (75--82) & 9.33 & +14.9 & 73\% & 17.4 & 8 \\
 & high & 80.2 (77--84) & 9.46 & +14.4 & 78\% & 17.2 & 44 \\
\midrule
\multirow{3}{*}{GPT-6.1 Sol} & low & 77.6 (74--81) & 9.34 & +15.7 & 72\% & 18.5 & 3 \\
 & medium & 78.6 (75--82) & 9.37 & +15.1 & 74\% & 17.8 & 10 \\
 & high & 78.4 (75--82) & 9.49 & +16.5 & 78\% & 18.3 & 42 \\
\midrule
\multirow{3}{*}{GPT-6 Luna} & low & 74.0 (70--78) & 8.73 & +13.3 & 41\% & 15.0 & 46 \\
 & medium & 77.2 (73--81) & 8.86 & +11.4 & 43\% & 13.9 & 120 \\
 & high & 76.4 (72--80) & 8.89 & +12.5 & 46\% & 16.7 & 161 \\
\midrule
\multirow{3}{*}{Claude Fable 5.1} & low & 82.3 (79--85) & 7.96 & -2.6 & 3\% & 3.3 & 0 \\
 & medium & 82.9 (79--86) & 7.99 & -2.9 & 3\% & 4.0 & 13 \\
 & high & 83.5 (80--86) & 8.06 & -2.8 & 2\% & 2.8 & 61 \\
\midrule
\multirow{3}{*}{Claude Opus 5.5} & low & 84.9 (81--88) & 7.84 & -6.5 & 2\% & 6.5 & 7 \\
 & medium & 85.3 (82--88) & 7.87 & -6.6 & 3\% & 6.6 & 61 \\
 & high & 85.7 (82--88) & 7.88 & -6.9 & 3\% & 6.9 & 78 \\
\midrule
\multirow{3}{*}{Gemini 4 Argon} & low & 85.7 (82--88) & 9.68 & +11.1 & 79\% & 11.4 & 0 \\
 & medium & 86.3 (83--89) & 9.68 & +10.5 & 78\% & 11.0 & 105 \\
 & high & 88.1 (85--91) & 9.76 & +9.5 & 82\% & 10.0 & 774 \\
\midrule
\multirow{3}{*}{Gemini 3.1 Pro} & low & 78.6 (75--82) & 9.33 & +14.7 & 65\% & 16.0 & 305 \\
 & medium & 79.8 (76--83) & 9.44 & +14.5 & 69\% & 16.7 & 419 \\
 & high & 81.8 (78--85) & 9.65 & +14.7 & 82\% & 16.8 & 772 \\
\midrule
\multirow{2}{*}{DeepSeek V4.1 Flash} & low & 77.7 (74--81) & 9.22 & +14.5 & 56\% & 16.5 & 445 \\
 & high & 77.4 (73--81) & 9.33 & +15.9 & 60\% & 17.1 & 634 \\
\midrule
\multirow{3}{*}{Qwen 3.8 Max} & low & 78.4 (75--82) & 8.84 & +10.0 & 23\% & 12.0 & 394 \\
 & medium & 84.2 (81--87) & 9.00 & +5.8 & 37\% & 7.8 & 2094 \\
 & high & 84.0 (81--87) & 9.03 & +6.3 & 37\% & 6.7 & 2033 \\
\midrule
\multirow{2}{*}{Kimi K3} & low & 77.4 (74--81) & 9.11 & +13.6 & 41\% & 15.1 & 218 \\
 & high & 76.0 (72--80) & 9.04 & +14.4 & 38\% & 15.1 & 356 \\
\bottomrule
\end{tabular}

\end{table*}

\begin{table*}[tp]
\centering
\caption{\textbf{\webshop: paired change for every effort contrast.} Success in percentage points; confidence on the 0--10 scale. Shaded bold cells are significant at $p<0.05$ in a paired $t$-test; $^{\dagger}$: still significant after Holm correction over the 26 contrasts per metric; $^{*}$: significant only before correction (see \cref{tab:effects}).}
\label{tab:supp_effects_ws}
\small
\setlength{\tabcolsep}{4pt}
\begin{tabular}{l>{\columncolor{accbg}}c>{\columncolor{accbg}}c>{\columncolor{accbg}}c!{\vrule width 1.2pt}>{\columncolor{confbg}}c>{\columncolor{confbg}}c>{\columncolor{confbg}}c}
\toprule
& \multicolumn{3}{c!{\vrule width 1.2pt}}{\cellcolor{accbg}\textbf{Success rate (pts)}} & \multicolumn{3}{c}{\cellcolor{confbg}\textbf{Stated confidence (0-10 scale)}} \\
Model & low $\rightarrow$ medium & medium $\rightarrow$ high & low $\rightarrow$ high & low $\rightarrow$ medium & medium $\rightarrow$ high & low $\rightarrow$ high \\
\midrule
GPT-6 Astra & -0.4 & \cellcolor{accsig}\textbf{+1.8$^{*}$} & +1.4 & \cellcolor{confsig}\textbf{+0.06$^{\dagger}$} & \cellcolor{confsig}\textbf{+0.13$^{\dagger}$} & \cellcolor{confsig}\textbf{+0.19$^{\dagger}$} \\
GPT-6.1 Sol & +1.0 & -0.2 & +0.8 & \cellcolor{confsig}\textbf{+0.04$^{*}$} & \cellcolor{confsig}\textbf{+0.11$^{\dagger}$} & \cellcolor{confsig}\textbf{+0.15$^{\dagger}$} \\
GPT-6 Luna & \cellcolor{accsig}\textbf{+3.2$^{*}$} & -0.8 & +2.4 & \cellcolor{confsig}\textbf{+0.13$^{\dagger}$} & +0.03 & \cellcolor{confsig}\textbf{+0.16$^{\dagger}$} \\
Claude Fable 5.1 & +0.6 & +0.6 & +1.2 & +0.03 & \cellcolor{confsig}\textbf{+0.07$^{\dagger}$} & \cellcolor{confsig}\textbf{+0.10$^{\dagger}$} \\
Claude Opus 5.5 & +0.4 & +0.4 & +0.8 & +0.03 & +0.01 & \cellcolor{confsig}\textbf{+0.04$^{*}$} \\
Gemini 4 Argon & +0.6 & \cellcolor{accsig}\textbf{+1.8$^{*}$} & \cellcolor{accsig}\textbf{+2.4$^{*}$} & +0.00 & \cellcolor{confsig}\textbf{+0.08$^{\dagger}$} & \cellcolor{confsig}\textbf{+0.08$^{\dagger}$} \\
Gemini 3.1 Pro & +1.2 & \cellcolor{accsig}\textbf{+2.0$^{*}$} & \cellcolor{accsig}\textbf{+3.2$^{*}$} & \cellcolor{confsig}\textbf{+0.10$^{\dagger}$} & \cellcolor{confsig}\textbf{+0.22$^{\dagger}$} & \cellcolor{confsig}\textbf{+0.32$^{\dagger}$} \\
DeepSeek V4.1 Flash & -- & -- & -0.2 & -- & -- & \cellcolor{confsig}\textbf{+0.11$^{\dagger}$} \\
Qwen 3.8 Max & \cellcolor{accsig}\textbf{+5.9$^{\dagger}$} & -0.4 & \cellcolor{accsig}\textbf{+5.5$^{\dagger}$} & \cellcolor{confsig}\textbf{+0.16$^{\dagger}$} & +0.03 & \cellcolor{confsig}\textbf{+0.19$^{\dagger}$} \\
Kimi K3 & -- & -- & -1.4 & -- & -- & \cellcolor{confsig}\textbf{-0.06$^{*}$} \\
\bottomrule
\end{tabular}

\end{table*}

\begin{table*}[tp]
\centering
\caption{\textbf{\webshop calibration by effort level.} ECE in points, Brier score, AUROC; metric blocks separated by heavy rules.}
\label{tab:supp_cal_ws}
\small
\setlength{\tabcolsep}{4pt}
\begin{tabular}{l>{\columncolor{ecebg}}c>{\columncolor{ecebg}}c>{\columncolor{ecebg}}c!{\vrule width 1.2pt}>{\columncolor{brierbg}}c>{\columncolor{brierbg}}c>{\columncolor{brierbg}}c!{\vrule width 1.2pt}>{\columncolor{aurocbg}}c>{\columncolor{aurocbg}}c>{\columncolor{aurocbg}}c}
\toprule
& \multicolumn{3}{c!{\vrule width 1.2pt}}{\cellcolor{ecebg}\textbf{ECE (pts)}} & \multicolumn{3}{c!{\vrule width 1.2pt}}{\cellcolor{brierbg}\textbf{Brier}} & \multicolumn{3}{c}{\cellcolor{aurocbg}\textbf{AUROC}} \\
Model & low & medium & high & low & medium & high & low & medium & high \\
\midrule
GPT-6 Astra & 18.0 & 17.4 & 17.2 & 0.203 & 0.200 & 0.187 & 0.55 & 0.56 & 0.56 \\
GPT-6.1 Sol & 18.5 & 17.8 & 18.3 & 0.208 & 0.202 & 0.198 & 0.56 & 0.54 & 0.57 \\
GPT-6 Luna & 15.0 & 13.9 & 16.7 & 0.208 & 0.193 & 0.205 & 0.61 & 0.59 & 0.58 \\
Claude Fable 5.1 & 3.3 & 4.0 & 2.8 & 0.136 & 0.135 & 0.133 & 0.70 & 0.69 & 0.67 \\
Claude Opus 5.5 & 6.5 & 6.6 & 6.9 & 0.129 & 0.126 & 0.122 & 0.69 & 0.69 & 0.70 \\
Gemini 4 Argon & 11.4 & 11.0 & 10.0 & 0.126 & 0.122 & 0.109 & 0.65 & 0.68 & 0.66 \\
Gemini 3.1 Pro & 16.0 & 16.7 & 16.8 & 0.184 & 0.182 & 0.177 & 0.66 & 0.63 & 0.54 \\
DeepSeek V4.1 Flash & 16.5 & -- & 17.1 & 0.201 & -- & 0.201 & 0.56 & -- & 0.58 \\
Qwen 3.8 Max & 12.0 & 7.8 & 6.7 & 0.183 & 0.132 & 0.131 & 0.59 & 0.64 & 0.65 \\
Kimi K3 & 15.1 & -- & 15.1 & 0.197 & -- & 0.200 & 0.56 & -- & 0.62 \\
\bottomrule
\end{tabular}

\end{table*}

\section{\sqid: full results}
\label{sec:supp_sq}

\textbf{Distractor selection.} To choose distractors, we choose items that are \texttt{Substitutes} first and then \texttt{Complements} only when \texttt{Substitutes} run out, so that every distractor is a product annotators judged close to, but short of, what the query asks for (so, \texttt{Irrelevant} products are never used). Also, a distractor whose title is identical to an \texttt{Exact} product of the same query is dropped as an inconsistently labeled duplicate listing. \texttt{Substitutes} make up 92\% of the total distractors and 424 of the 500 tasks contain only \texttt{Substitutes}, usually the same product in a different size, color, pack count or compatible model.

\begin{figure*}[tp]
\centering
\includegraphics[width=0.85\textwidth]{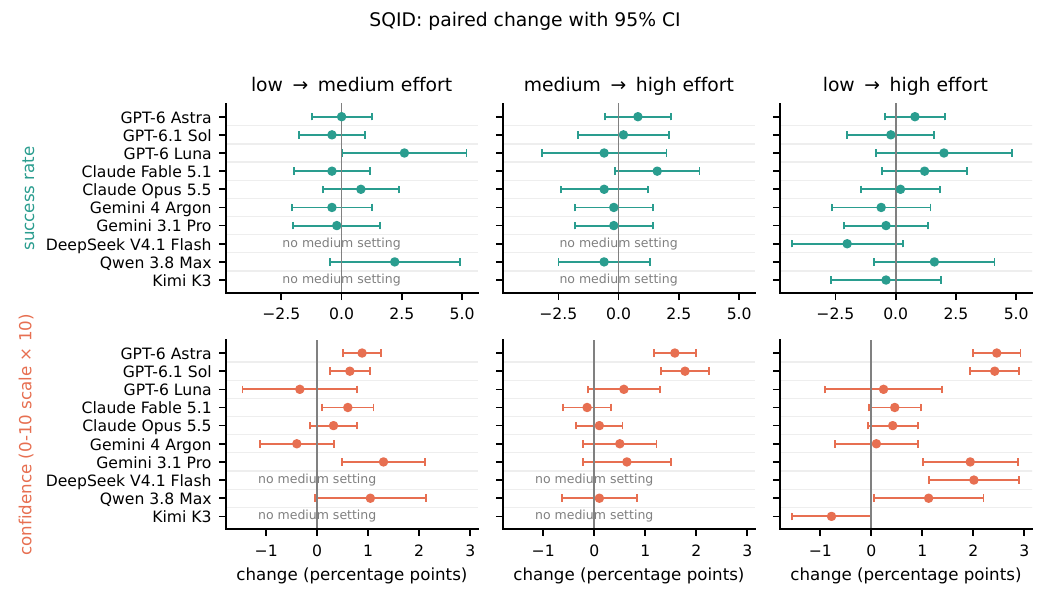}
\caption{\textbf{\sqid: paired within-task change with 95\% intervals} for success (top) and confidence (bottom) under each effort contrast. The only significant success rate change is GPT-6 Luna on the low$\rightarrow$medium case; confidence rises for five models in the low$\rightarrow$high case, four models in the low$\rightarrow$medium case, and two in the medium$\rightarrow$high case.}
\label{fig:supp_effects_sq}
\end{figure*}

\begin{figure*}[tp]
\centering
\includegraphics[width=\textwidth]{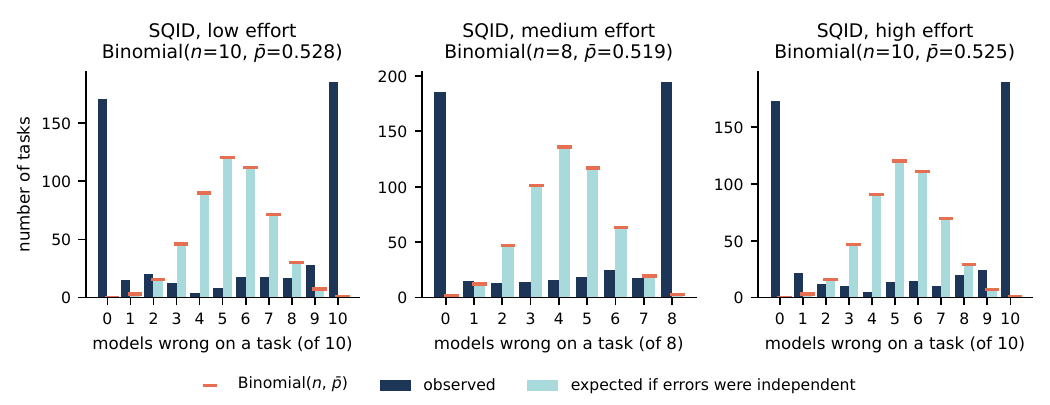}
\caption{\textbf{\sqid: number of models wrong per task} at each effort level, observed versus the independence null. About 38\% of tasks are failed by every model and 34-37\% solved by every model; independence predicts under 1\% for each.}
\label{fig:supp_conc_sq}
\end{figure*}

This section repeats every \webshop figure and table of the main paper for \sqid (500 tasks, 13,990 valid decisions): per-condition results (\cref{tab:supp_glance_sq}), the effort grid (\cref{fig:supp_grid}, middle row), all three paired effort contrasts (\cref{fig:supp_effects_sq}, \cref{tab:supp_effects_sq}), calibration by effort level (\cref{tab:supp_cal_sq}), the reliability diagram (\cref{fig:supp_rel_sqwa}, left), confidence when right versus wrong (\cref{fig:supp_rw_sqwa}, left) and error concentration at every effort level (\cref{fig:supp_conc_sq}, \cref{tab:supp_conc_sq}).

\begin{table*}[tp]
\centering
\caption{\textbf{\sqid results per condition.} Columns as in \cref{tab:supp_glance_ws}.}
\label{tab:supp_glance_sq}
\small
\begin{tabular}{llcccccc}
\toprule
Model & Effort & Success (95\% CI) & Conf. & Overconf. & 10/10 & ECE & Tokens \\
\midrule
\multirow{3}{*}{GPT-6 Astra} & low & 46.6 (42--51) & 7.75 & +30.9 & 41\% & 33.0 & 10 \\
 & medium & 46.6 (42--51) & 7.84 & +31.8 & 43\% & 32.5 & 22 \\
 & high & 47.4 (43--52) & 8.00 & +32.6 & 47\% & 33.8 & 77 \\
\midrule
\multirow{3}{*}{GPT-6.1 Sol} & low & 47.2 (43--52) & 7.86 & +31.4 & 42\% & 33.4 & 13 \\
 & medium & 46.8 (42--51) & 7.93 & +32.5 & 43\% & 33.2 & 25 \\
 & high & 47.0 (43--51) & 8.11 & +34.1 & 47\% & 35.1 & 81 \\
\midrule
\multirow{3}{*}{GPT-6 Luna} & low & 43.7 (39--48) & 7.46 & +30.9 & 24\% & 32.4 & 41 \\
 & medium & 46.4 (42--51) & 7.43 & +27.9 & 23\% & 28.3 & 145 \\
 & high & 45.7 (41--50) & 7.48 & +29.2 & 23\% & 30.4 & 189 \\
\midrule
\multirow{3}{*}{Claude Fable 5.1} & low & 48.8 (44--53) & 6.11 & +12.3 & 3\% & 12.8 & 3 \\
 & medium & 48.4 (44--53) & 6.17 & +13.3 & 3\% & 14.3 & 46 \\
 & high & 50.0 (46--54) & 6.16 & +11.6 & 3\% & 12.6 & 128 \\
\midrule
\multirow{3}{*}{Claude Opus 5.5} & low & 48.6 (44--53) & 5.94 & +10.8 & 0\% & 13.7 & 43 \\
 & medium & 49.4 (45--54) & 5.98 & +10.4 & 1\% & 11.8 & 111 \\
 & high & 48.8 (44--53) & 5.99 & +11.1 & 1\% & 11.8 & 127 \\
\midrule
\multirow{3}{*}{Gemini 4 Argon} & low & 50.6 (46--55) & 8.02 & +29.6 & 40\% & 29.7 & 0 \\
 & medium & 50.2 (46--55) & 7.98 & +29.6 & 40\% & 30.1 & 134 \\
 & high & 50.1 (46--54) & 8.03 & +30.2 & 38\% & 30.5 & 924 \\
\midrule
\multirow{3}{*}{Gemini 3.1 Pro} & low & 48.4 (44--53) & 7.72 & +28.8 & 31\% & 32.0 & 287 \\
 & medium & 48.2 (44--53) & 7.85 & +30.3 & 36\% & 31.2 & 399 \\
 & high & 48.0 (44--52) & 7.91 & +31.1 & 41\% & 32.8 & 877 \\
\midrule
\multirow{2}{*}{DeepSeek V4.1 Flash} & low & 47.1 (43--51) & 7.86 & +31.5 & 26\% & 32.8 & 567 \\
 & high & 45.0 (41--49) & 8.07 & +35.7 & 31\% & 35.9 & 929 \\
\midrule
\multirow{3}{*}{Qwen 3.8 Max} & low & 46.3 (42--51) & 7.34 & +27.2 & 16\% & 28.1 & 488 \\
 & medium & 48.5 (44--53) & 7.45 & +26.0 & 20\% & 27.4 & 3901 \\
 & high & 48.0 (44--52) & 7.46 & +26.6 & 21\% & 27.4 & 3926 \\
\midrule
\multirow{2}{*}{Kimi K3} & low & 46.2 (42--51) & 7.88 & +32.6 & 22\% & 32.7 & 206 \\
 & high & 45.8 (41--50) & 7.80 & +32.2 & 17\% & 32.2 & 319 \\
\bottomrule
\end{tabular}

\end{table*}

\begin{table*}[tp]
\centering
\caption{\textbf{\sqid: paired change for every effort contrast.} Success in percentage points; confidence on the 0--10 scale. Shaded bold cells are significant at $p<0.05$ in a paired $t$-test; $^{\dagger}$: still significant after Holm correction over the 26 contrasts per metric; $^{*}$: significant only before correction (see \cref{tab:effects}).}
\label{tab:supp_effects_sq}
\small
\setlength{\tabcolsep}{4pt}
\begin{tabular}{l>{\columncolor{accbg}}c>{\columncolor{accbg}}c>{\columncolor{accbg}}c!{\vrule width 1.2pt}>{\columncolor{confbg}}c>{\columncolor{confbg}}c>{\columncolor{confbg}}c}
\toprule
& \multicolumn{3}{c!{\vrule width 1.2pt}}{\cellcolor{accbg}\textbf{Success rate (pts)}} & \multicolumn{3}{c}{\cellcolor{confbg}\textbf{Stated confidence (0-10 scale)}} \\
Model & low $\rightarrow$ medium & medium $\rightarrow$ high & low $\rightarrow$ high & low $\rightarrow$ medium & medium $\rightarrow$ high & low $\rightarrow$ high \\
\midrule
GPT-6 Astra & +0.0 & +0.8 & +0.8 & \cellcolor{confsig}\textbf{+0.09$^{\dagger}$} & \cellcolor{confsig}\textbf{+0.16$^{\dagger}$} & \cellcolor{confsig}\textbf{+0.25$^{\dagger}$} \\
GPT-6.1 Sol & -0.4 & +0.2 & -0.2 & \cellcolor{confsig}\textbf{+0.06$^{\dagger}$} & \cellcolor{confsig}\textbf{+0.18$^{\dagger}$} & \cellcolor{confsig}\textbf{+0.24$^{\dagger}$} \\
GPT-6 Luna & \cellcolor{accsig}\textbf{+2.6$^{*}$} & -0.6 & +2.0 & -0.03 & +0.06 & +0.02 \\
Claude Fable 5.1 & -0.4 & +1.6 & +1.2 & \cellcolor{confsig}\textbf{+0.06$^{*}$} & -0.01 & +0.05 \\
Claude Opus 5.5 & +0.8 & -0.6 & +0.2 & +0.03 & +0.01 & +0.04 \\
Gemini 4 Argon & -0.4 & -0.2 & -0.6 & -0.04 & +0.05 & +0.01 \\
Gemini 3.1 Pro & -0.2 & -0.2 & -0.4 & \cellcolor{confsig}\textbf{+0.13$^{\dagger}$} & +0.06 & \cellcolor{confsig}\textbf{+0.19$^{\dagger}$} \\
DeepSeek V4.1 Flash & -- & -- & -2.0 & -- & -- & \cellcolor{confsig}\textbf{+0.20$^{\dagger}$} \\
Qwen 3.8 Max & +2.2 & -0.6 & +1.6 & +0.10 & +0.01 & \cellcolor{confsig}\textbf{+0.11$^{*}$} \\
Kimi K3 & -- & -- & -0.4 & -- & -- & -0.08 \\
\bottomrule
\end{tabular}

\end{table*}

\begin{table*}[tp]
\centering
\caption{\textbf{\sqid calibration by effort level.} ECE in points, Brier score, AUROC; metric blocks separated by heavy rules.}
\label{tab:supp_cal_sq}
\small
\setlength{\tabcolsep}{4pt}
\begin{tabular}{l>{\columncolor{ecebg}}c>{\columncolor{ecebg}}c>{\columncolor{ecebg}}c!{\vrule width 1.2pt}>{\columncolor{brierbg}}c>{\columncolor{brierbg}}c>{\columncolor{brierbg}}c!{\vrule width 1.2pt}>{\columncolor{aurocbg}}c>{\columncolor{aurocbg}}c>{\columncolor{aurocbg}}c}
\toprule
& \multicolumn{3}{c!{\vrule width 1.2pt}}{\cellcolor{ecebg}\textbf{ECE (pts)}} & \multicolumn{3}{c!{\vrule width 1.2pt}}{\cellcolor{brierbg}\textbf{Brier}} & \multicolumn{3}{c}{\cellcolor{aurocbg}\textbf{AUROC}} \\
Model & low & medium & high & low & medium & high & low & medium & high \\
\midrule
GPT-6 Astra & 33.0 & 32.5 & 33.8 & 0.349 & 0.349 & 0.352 & 0.66 & 0.67 & 0.67 \\
GPT-6.1 Sol & 33.4 & 33.2 & 35.1 & 0.354 & 0.347 & 0.361 & 0.65 & 0.68 & 0.67 \\
GPT-6 Luna & 32.4 & 28.3 & 30.4 & 0.344 & 0.310 & 0.320 & 0.67 & 0.70 & 0.70 \\
Claude Fable 5.1 & 12.8 & 14.3 & 12.6 & 0.223 & 0.226 & 0.227 & 0.74 & 0.74 & 0.73 \\
Claude Opus 5.5 & 13.7 & 11.8 & 11.8 & 0.229 & 0.221 & 0.224 & 0.72 & 0.73 & 0.73 \\
Gemini 4 Argon & 29.7 & 30.1 & 30.5 & 0.318 & 0.323 & 0.319 & 0.69 & 0.69 & 0.70 \\
Gemini 3.1 Pro & 32.0 & 31.2 & 32.8 & 0.336 & 0.335 & 0.334 & 0.69 & 0.69 & 0.70 \\
DeepSeek V4.1 Flash & 32.8 & -- & 35.9 & 0.335 & -- & 0.353 & 0.71 & -- & 0.74 \\
Qwen 3.8 Max & 28.1 & 27.4 & 27.4 & 0.304 & 0.282 & 0.283 & 0.72 & 0.75 & 0.75 \\
Kimi K3 & 32.7 & -- & 32.2 & 0.336 & -- & 0.326 & 0.70 & -- & 0.72 \\
\bottomrule
\end{tabular}

\end{table*}

\begin{table*}[p]
\centering
\caption{\textbf{\sqid error concentration by effort level.} Columns as in the main-paper concentration table.}
\label{tab:supp_conc_sq}
\footnotesize
\setlength{\tabcolsep}{3pt}
\resizebox{\columnwidth}{!}{\begin{tabular}{lcccccc}
\toprule
Effort & $n$, $\bar p$ & All wrong & All right & Var ratio & $\rho$ & KL div. \\
& & obs / indep & obs / indep & & & \\
\midrule
low & 10, 0.528 & 185 / 0.8 & 171 / 0.3 & 8.1 & 0.79 & 4.12 \\
medium & 8, 0.519 & 194 / 2.6 & 185 / 1.4 & 6.6 & 0.80 & 3.21 \\
high & 10, 0.525 & 189 / 0.8 & 173 / 0.3 & 8.3 & 0.81 & 4.24 \\
\bottomrule
\end{tabular}
}
\end{table*}

\section{\wands: full results}
\label{sec:supp_wa}

\begin{figure*}[tp]
\centering
\includegraphics[width=0.85\textwidth]{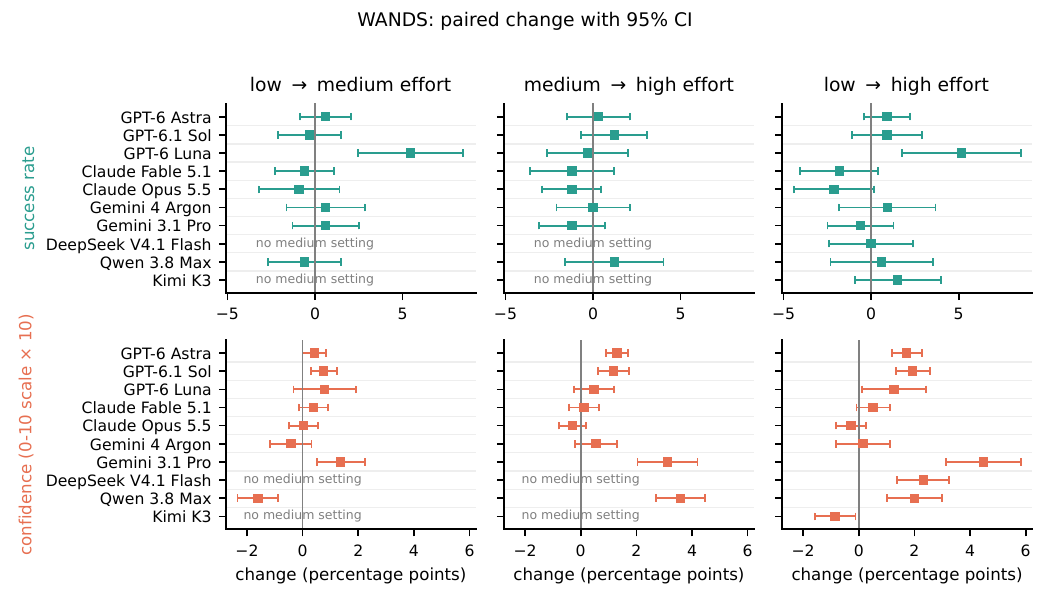}
\caption{\textbf{\wands: paired within-task change with 95\% intervals} under each effort contrast. GPT-6 Luna is the only model with a significant success gain, entirely in the low$\rightarrow$medium step.}
\label{fig:supp_effects_wa}
\end{figure*}

\begin{figure*}[tp]
\centering
\includegraphics[width=\textwidth]{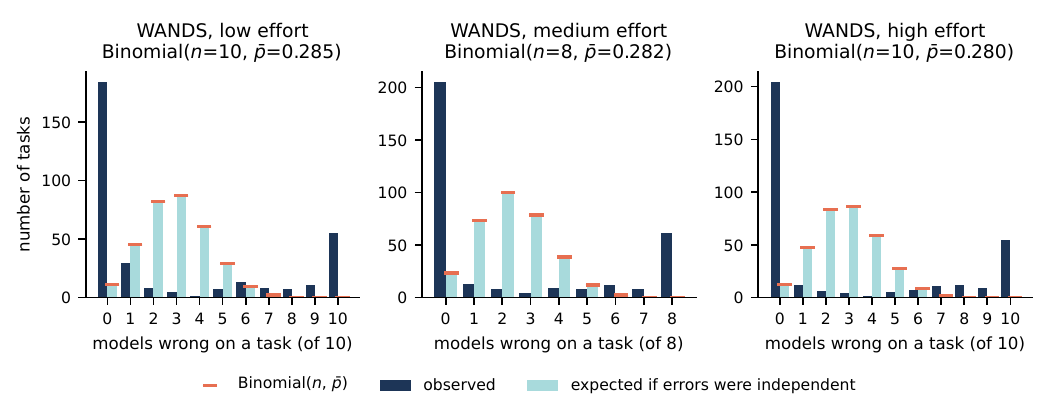}
\caption{\textbf{\wands: number of models wrong per task} at each effort level, observed versus the independence null.}
\label{fig:supp_conc_wa}
\end{figure*}

The same material for \wands (330 tasks, 9,234 valid decisions): \cref{tab:supp_glance_wa}, \cref{fig:supp_grid} (bottom row), \cref{fig:supp_effects_wa}, \cref{tab:supp_effects_wa}, \cref{tab:supp_cal_wa}, \cref{fig:supp_rel_sqwa} (right), \cref{fig:supp_rw_sqwa} (right), \cref{fig:supp_conc_wa} and \cref{tab:supp_conc_wa}.

\begin{table*}[tp]
\centering
\caption{\textbf{\wands results per condition.} Columns as in \cref{tab:supp_glance_ws}.}
\label{tab:supp_glance_wa}
\small
\begin{tabular}{llcccccc}
\toprule
Model & Effort & Success (95\% CI) & Conf. & Overconf. & 10/10 & ECE & Tokens \\
\midrule
\multirow{3}{*}{GPT-6 Astra} & low & 71.5 (66--76) & 8.98 & +18.3 & 61\% & 18.7 & 6 \\
 & medium & 72.1 (67--77) & 9.02 & +18.1 & 63\% & 18.7 & 14 \\
 & high & 72.4 (67--77) & 9.15 & +19.1 & 68\% & 19.8 & 54 \\
\midrule
\multirow{3}{*}{GPT-6.1 Sol} & low & 71.5 (66--76) & 9.06 & +19.1 & 63\% & 19.4 & 7 \\
 & medium & 71.2 (66--76) & 9.13 & +20.1 & 65\% & 20.1 & 15 \\
 & high & 72.4 (67--77) & 9.25 & +20.1 & 69\% & 20.1 & 49 \\
\midrule
\multirow{3}{*}{GPT-6 Luna} & low & 65.8 (60--71) & 8.22 & +16.4 & 30\% & 18.8 & 22 \\
 & medium & 71.2 (66--76) & 8.30 & +11.8 & 35\% & 12.8 & 97 \\
 & high & 70.9 (66--76) & 8.35 & +12.5 & 37\% & 12.7 & 135 \\
\midrule
\multirow{3}{*}{Claude Fable 5.1} & low & 72.7 (68--77) & 7.23 & -0.4 & 6\% & 3.6 & 0 \\
 & medium & 72.1 (67--77) & 7.27 & +0.6 & 5\% & 4.8 & 31 \\
 & high & 70.9 (66--76) & 7.28 & +1.9 & 6\% & 6.5 & 105 \\
\midrule
\multirow{3}{*}{Claude Opus 5.5} & low & 72.4 (67--77) & 7.00 & -2.5 & 2\% & 3.1 & 20 \\
 & medium & 71.5 (66--76) & 7.00 & -1.5 & 3\% & 7.2 & 65 \\
 & high & 70.3 (65--75) & 6.97 & -0.6 & 3\% & 5.7 & 79 \\
\midrule
\multirow{3}{*}{Gemini 4 Argon} & low & 72.7 (68--77) & 9.12 & +18.4 & 64\% & 19.5 & 0 \\
 & medium & 73.3 (68--78) & 9.07 & +17.4 & 60\% & 17.9 & 106 \\
 & high & 73.7 (69--78) & 9.17 & +18.0 & 59\% & 18.3 & 715 \\
\midrule
\multirow{3}{*}{Gemini 3.1 Pro} & low & 72.1 (67--77) & 8.63 & +14.2 & 45\% & 14.8 & 267 \\
 & medium & 72.7 (68--77) & 8.76 & +14.9 & 48\% & 15.3 & 345 \\
 & high & 71.5 (66--76) & 9.08 & +19.2 & 59\% & 19.2 & 600 \\
\midrule
\multirow{2}{*}{DeepSeek V4.1 Flash} & low & 71.1 (66--76) & 8.75 & +16.4 & 40\% & 17.0 & 329 \\
 & high & 70.8 (66--75) & 8.98 & +19.0 & 48\% & 20.0 & 623 \\
\midrule
\multirow{3}{*}{Qwen 3.8 Max} & low & 71.8 (67--76) & 8.17 & +9.9 & 25\% & 11.4 & 464 \\
 & medium & 71.1 (66--76) & 8.01 & +9.0 & 22\% & 11.8 & 589 \\
 & high & 72.4 (67--77) & 8.37 & +11.3 & 35\% & 12.1 & 788 \\
\midrule
\multirow{2}{*}{Kimi K3} & low & 71.2 (66--76) & 8.75 & +16.3 & 38\% & 16.7 & 153 \\
 & high & 72.7 (68--77) & 8.67 & +14.0 & 34\% & 14.2 & 212 \\
\bottomrule
\end{tabular}

\end{table*}

\begin{table*}[tp]
\centering
\caption{\textbf{\wands: paired change for every effort contrast.} Format as in \cref{tab:supp_effects_sq}.}
\label{tab:supp_effects_wa}
\small
\setlength{\tabcolsep}{4pt}
\begin{tabular}{l>{\columncolor{accbg}}c>{\columncolor{accbg}}c>{\columncolor{accbg}}c!{\vrule width 1.2pt}>{\columncolor{confbg}}c>{\columncolor{confbg}}c>{\columncolor{confbg}}c}
\toprule
& \multicolumn{3}{c!{\vrule width 1.2pt}}{\cellcolor{accbg}\textbf{Success rate (pts)}} & \multicolumn{3}{c}{\cellcolor{confbg}\textbf{Stated confidence (0-10 scale)}} \\
Model & low $\rightarrow$ medium & medium $\rightarrow$ high & low $\rightarrow$ high & low $\rightarrow$ medium & medium $\rightarrow$ high & low $\rightarrow$ high \\
\midrule
GPT-6 Astra & +0.6 & +0.3 & +0.9 & +0.04 & \cellcolor{confsig}\textbf{+0.13$^{\dagger}$} & \cellcolor{confsig}\textbf{+0.17$^{\dagger}$} \\
GPT-6.1 Sol & -0.3 & +1.2 & +0.9 & \cellcolor{confsig}\textbf{+0.08$^{\dagger}$} & \cellcolor{confsig}\textbf{+0.12$^{\dagger}$} & \cellcolor{confsig}\textbf{+0.19$^{\dagger}$} \\
GPT-6 Luna & \cellcolor{accsig}\textbf{+5.5$^{\dagger}$} & -0.3 & \cellcolor{accsig}\textbf{+5.2$^{*}$} & +0.08 & +0.05 & \cellcolor{confsig}\textbf{+0.13$^{*}$} \\
Claude Fable 5.1 & -0.6 & -1.2 & -1.8 & +0.04 & +0.01 & +0.05 \\
Claude Opus 5.5 & -0.9 & -1.2 & -2.1 & +0.00 & -0.03 & -0.03 \\
Gemini 4 Argon & +0.6 & +0.0 & +0.9 & -0.04 & +0.06 & +0.02 \\
Gemini 3.1 Pro & +0.6 & -1.2 & -0.6 & \cellcolor{confsig}\textbf{+0.14$^{\dagger}$} & \cellcolor{confsig}\textbf{+0.31$^{\dagger}$} & \cellcolor{confsig}\textbf{+0.45$^{\dagger}$} \\
DeepSeek V4.1 Flash & -- & -- & +0.0 & -- & -- & \cellcolor{confsig}\textbf{+0.23$^{\dagger}$} \\
Qwen 3.8 Max & -0.6 & +1.2 & +0.6 & \cellcolor{confsig}\textbf{-0.16$^{\dagger}$} & \cellcolor{confsig}\textbf{+0.36$^{\dagger}$} & \cellcolor{confsig}\textbf{+0.20$^{\dagger}$} \\
Kimi K3 & -- & -- & +1.5 & -- & -- & \cellcolor{confsig}\textbf{-0.08$^{*}$} \\
\bottomrule
\end{tabular}

\end{table*}

\begin{table*}[tp]
\centering
\caption{\textbf{\wands calibration by effort level.}}
\label{tab:supp_cal_wa}
\small
\setlength{\tabcolsep}{4pt}
\begin{tabular}{l>{\columncolor{ecebg}}c>{\columncolor{ecebg}}c>{\columncolor{ecebg}}c!{\vrule width 1.2pt}>{\columncolor{brierbg}}c>{\columncolor{brierbg}}c>{\columncolor{brierbg}}c!{\vrule width 1.2pt}>{\columncolor{aurocbg}}c>{\columncolor{aurocbg}}c>{\columncolor{aurocbg}}c}
\toprule
& \multicolumn{3}{c!{\vrule width 1.2pt}}{\cellcolor{ecebg}\textbf{ECE (pts)}} & \multicolumn{3}{c!{\vrule width 1.2pt}}{\cellcolor{brierbg}\textbf{Brier}} & \multicolumn{3}{c}{\cellcolor{aurocbg}\textbf{AUROC}} \\
Model & low & medium & high & low & medium & high & low & medium & high \\
\midrule
GPT-6 Astra & 18.7 & 18.7 & 19.8 & 0.213 & 0.214 & 0.215 & 0.68 & 0.68 & 0.67 \\
GPT-6.1 Sol & 19.4 & 20.1 & 20.1 & 0.215 & 0.221 & 0.214 & 0.69 & 0.68 & 0.69 \\
GPT-6 Luna & 18.8 & 12.8 & 12.7 & 0.224 & 0.186 & 0.184 & 0.75 & 0.76 & 0.77 \\
Claude Fable 5.1 & 3.6 & 4.8 & 6.5 & 0.153 & 0.157 & 0.156 & 0.79 & 0.79 & 0.80 \\
Claude Opus 5.5 & 3.1 & 7.2 & 5.7 & 0.155 & 0.156 & 0.154 & 0.79 & 0.81 & 0.82 \\
Gemini 4 Argon & 19.5 & 17.9 & 18.3 & 0.206 & 0.193 & 0.189 & 0.73 & 0.74 & 0.77 \\
Gemini 3.1 Pro & 14.8 & 15.3 & 19.2 & 0.182 & 0.180 & 0.200 & 0.77 & 0.76 & 0.77 \\
DeepSeek V4.1 Flash & 17.0 & -- & 20.0 & 0.197 & -- & 0.212 & 0.78 & -- & 0.77 \\
Qwen 3.8 Max & 11.4 & 11.8 & 12.1 & 0.182 & 0.181 & 0.175 & 0.74 & 0.76 & 0.77 \\
Kimi K3 & 16.7 & -- & 14.2 & 0.197 & -- & 0.178 & 0.75 & -- & 0.77 \\
\bottomrule
\end{tabular}

\end{table*}

\begin{table*}[tp]
\centering
\caption{\textbf{\wands error concentration by effort level.}}
\label{tab:supp_conc_wa}
\footnotesize
\setlength{\tabcolsep}{3pt}
\resizebox{\columnwidth}{!}{\begin{tabular}{lcccccc}
\toprule
Effort & $n$, $\bar p$ & All wrong & All right & Var ratio & $\rho$ & KL div. \\
& & obs / indep & obs / indep & & & \\
\midrule
low & 10, 0.285 & 55 / 0.0 & 184 / 11.4 & 7.9 & 0.77 & 3.48 \\
medium & 8, 0.282 & 62 / 0.0 & 205 / 23.3 & 6.7 & 0.81 & 2.88 \\
high & 10, 0.280 & 55 / 0.0 & 204 / 12.3 & 8.3 & 0.81 & 3.76 \\
\bottomrule
\end{tabular}
}
\end{table*}

\section{Cross-dataset views}
\label{sec:supp_cross}

\begin{figure*}[tp]
\centering
\includegraphics[width=0.9\textwidth]{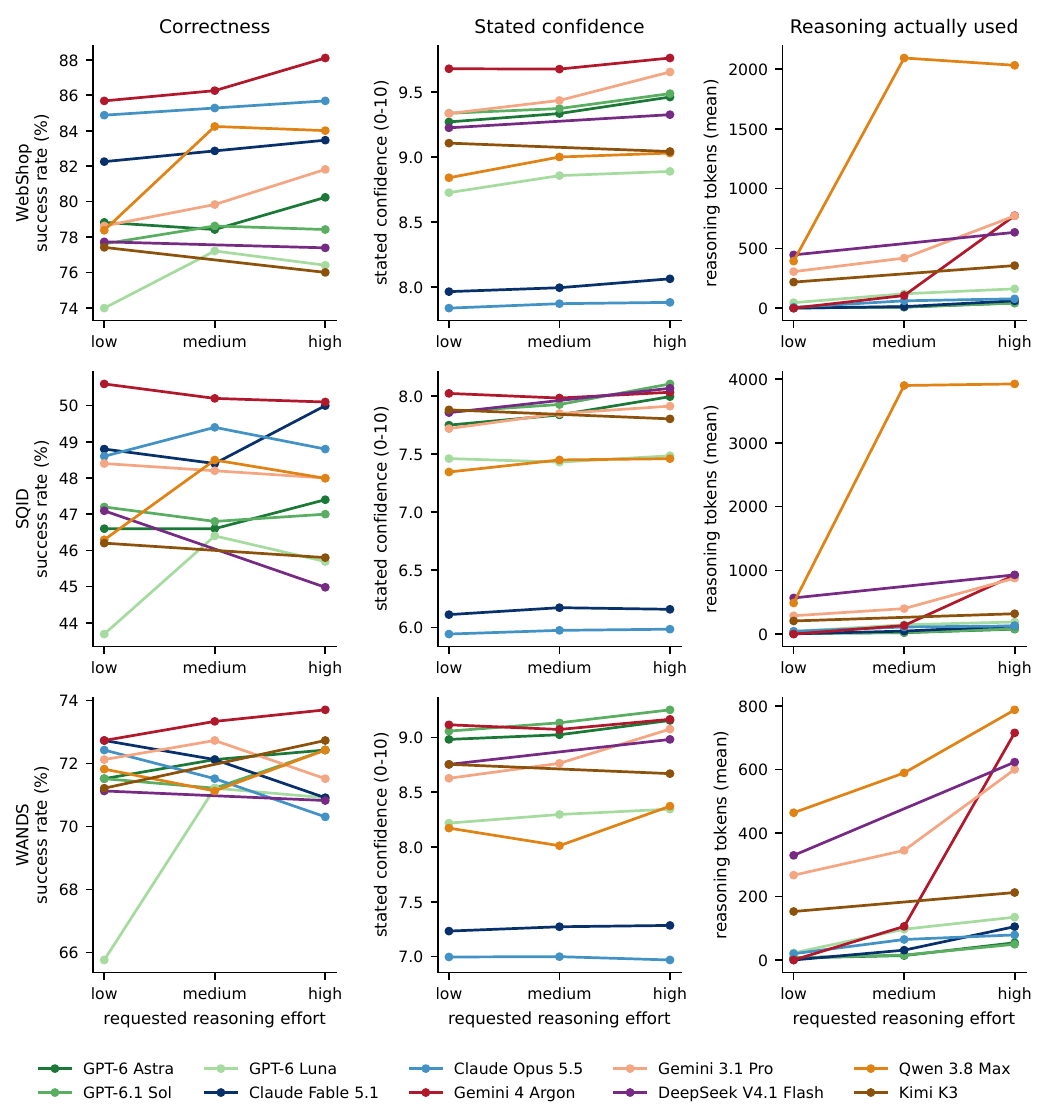}
\caption{\textbf{Success, confidence and reasoning tokens by requested effort on all three datasets.} Rows: \webshop, \sqid, \wands. All 28 \webshop conditions lie between 74.0\% and 88.1\% success, all \sqid conditions between 43.7\% and 50.6\%, and all \wands conditions between 65.8\% and 73.7\%. \sqid consistently has the lowest accuracy.}
\label{fig:supp_grid}
\end{figure*}

\begin{figure*}[tp]
\centering
\begin{minipage}[t]{0.49\textwidth}
\centering
\includegraphics[width=0.95\linewidth]{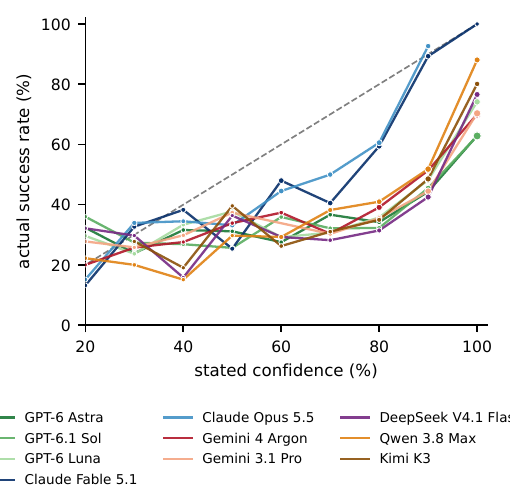}
\end{minipage}\hfill
\begin{minipage}[t]{0.49\textwidth}
\centering
\includegraphics[width=0.95\linewidth]{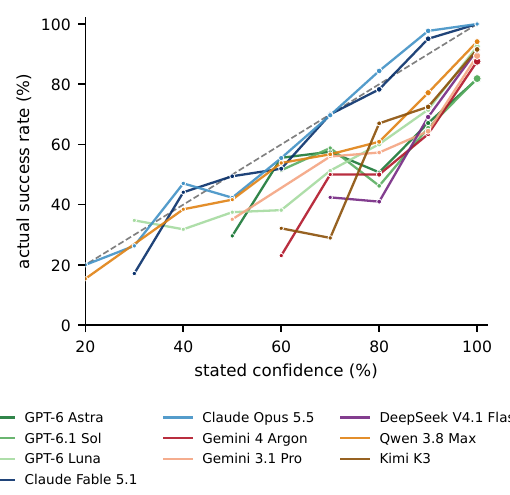}
\end{minipage}
\caption{\textbf{Reliability diagrams for \sqid (left) and \wands (right).} On \sqid every model's success rate lies below the diagonal at most confidence levels. Curves touch or cross it only at $20–30\%$ and, for the Claude models alone, at $90–100\%$. The Claude models are most often the closest to the diagonal in both cases.}
\label{fig:supp_rel_sqwa}
\end{figure*}

\begin{figure*}[tp]
\centering
\begin{minipage}[t]{0.49\textwidth}
\centering
\includegraphics[width=0.95\linewidth]{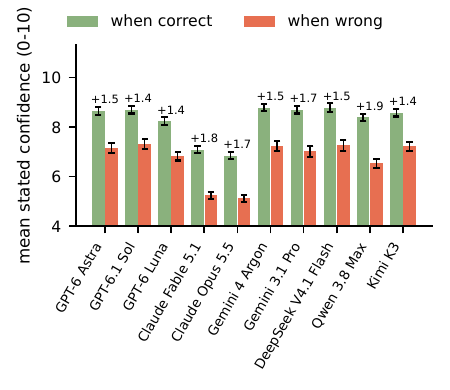}
\end{minipage}\hfill
\begin{minipage}[t]{0.49\textwidth}
\centering
\includegraphics[width=0.95\linewidth]{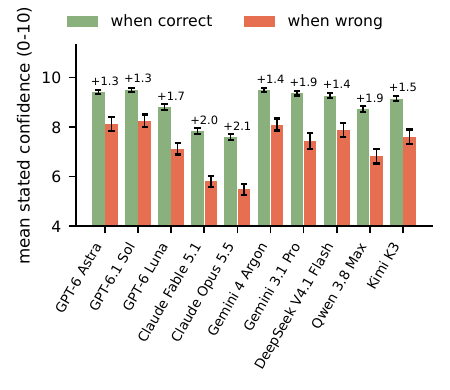}
\end{minipage}
\caption{\textbf{Mean stated confidence on correct versus wrong answers} for \sqid (left) and \wands (right), with 95\% intervals. The drops are larger than on \webshop (1.3--2.1 against 0.3--0.8) but mean confidence on wrong answers still exceeds 6.5/10 for every non-Claude model. The Claude models drop on \wands by 2.02 and 2.10 against 1.26--1.92, and on \sqid by 1.84 and 1.72 against 1.37--1.85 (Qwen 3.8 Max is marginally first).}
\label{fig:supp_rw_sqwa}
\end{figure*}

\begin{figure*}[tp]
\centering
\includegraphics[width=\textwidth]{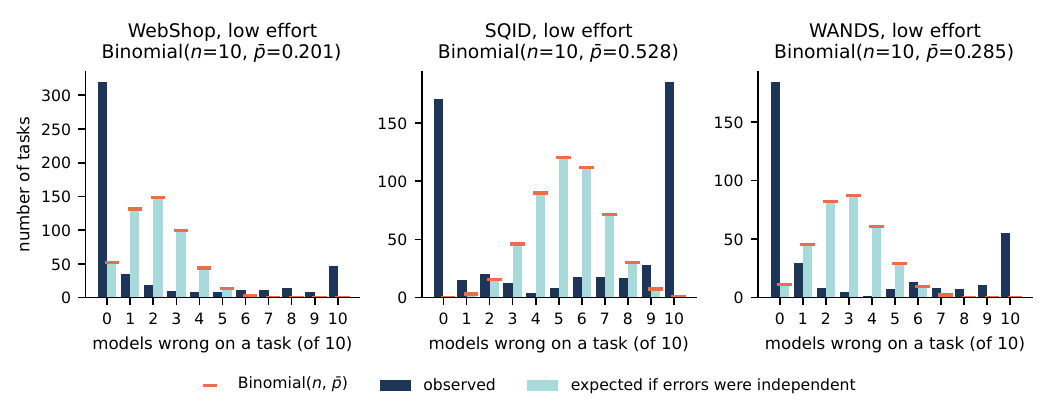}
\caption{\textbf{Low-effort error concentration on the three datasets}, observed versus independence.}
\label{fig:supp_conc_low}
\end{figure*}

\Cref{fig:supp_grid} stacks the effort grids of the three datasets. Within a dataset, models are spread over 4--11 points, and the ordering is only partly stable: success ranks agree closely between \webshop and \sqid (Spearman $\rho=0.96$ over the ten models) but weakly between either and \wands ($0.47$ and $0.50$), where text-only furniture queries change which models do well. \Cref{fig:supp_rel_sqwa,fig:supp_rw_sqwa} give the reliability diagrams and right-versus-wrong confidence for \sqid and \wands; \cref{fig:supp_conc_low} shows the low-effort concentration histograms of the three datasets side by side.

\section{Confidence when right versus wrong}
\label{sec:rightwrong}

\Cref{fig:rightwrong} splits \webshop confidence by outcome; \cref{fig:supp_rw_sqwa} does the same for \sqid and \wands. Every model on every dataset rates its correct answers higher than its wrong ones, which is the discrimination AUROC captures, but the size of the drop is model-dependent and the Claude models are consistently near the top. More important for a downstream user is the level. The mean confidence on \emph{wrong} answers is much lower for Claude models than for non-Claude ones, and the 95\% intervals are narrow relative to these gaps.

\section{Prompt}
\label{sec:supp_prompt}

Every request used the following system prompt and answer instructions. The user message consisted of the customer instruction, the line ``\emph{Here are 10 products. Each lists its title, price, features, description, and selectable options}'' (``\emph{, followed by its main image}'' when images were sent), the ten product blocks and then the answer instructions. In SQID and WANDS, there is no price, so the price line in the previous prompt is omitted in those datasets.

\begin{quote}\small
\textbf{System.} You are a shopping assistant. A customer gives you a shopping instruction, and you must buy the single product, with the right options, that best satisfies it.

\textbf{Answer instructions.} Choose exactly one product. For each of its option types, choose exactly one of the listed values. Then rate how confident you are that your chosen product and options satisfy every requirement in the customer instruction, as an integer from 0 (not at all confident) to 10 (completely confident).

\raggedright
Respond with only a JSON object in this format:
\texttt{\{"product": <product number>, "options": \{"<option type>": "<value>"\}, "confidence": <integer 0-10>\}}
Use \texttt{"options": \{\}} if the chosen product has no options.
\end{quote}

Each product block had the form ``\texttt{--- Product $k$ ---}'', then \texttt{Title}, \texttt{Price} (\webshop only), \texttt{Features} as bullets, \texttt{Description} (truncated at 1,500 characters) and \texttt{Options} (``none'' for \sqid and \wands), followed by the product image where available.

\begin{table*}[tp]
\centering
\caption{\textbf{Example \webshop tasks.} Instruction text is shown verbatim; product titles are abridged. ``Why distractors fail'' summarizes the simulator's verdict on the nine non-target candidates in that task. Construction rules are given in \cref{sec:tasks}.}
\label{tab:supp_examples}
\small
\setlength{\tabcolsep}{5pt}
\begin{tabular}{>{\raggedright\arraybackslash}p{0.34\textwidth}>{\raggedright\arraybackslash}p{0.24\textwidth}>{\raggedright\arraybackslash}p{0.34\textwidth}}
\toprule
Instruction & Correct product & Why distractors fail \\
\midrule
``i need an oval soft rug that is easy to clean. also, it should be in eggplant purple color, and price lower than 270.00 dollars'' &
Unique Loom Solo Solid Shag area rug (\$254.84), with \emph{shape: oval} and \emph{color: eggplant purple} selected &
All nine are shag or area rugs under \$270 from the same search, but none offers \emph{eggplant purple} as a color option (one offers purple but not an oval shape). \\
\addlinespace
``i want to get a super soft blue fleece throw that's 50 inches by 63 inches, and price lower than 50.00 dollars'' &
HILHORA flannel fleece throw (\$16.99), with \emph{size: 50$''\times$63$''$} and \emph{color: blue} selected &
Six throws do not offer blue; two are not fleece (faux fur, woven); one is not described as super soft; none other than the target offers the 50$''\times$63$''$ size. \\
\addlinespace
``i just love the matilde vicenzi brand macaroons and i want to try the amaretto d'italia macaroons flavor. the quality of the ingredients is my requirement, if you find it let me know, and price lower than 50.00 dollars'' &
Amaretto d'Italia Macaroons by Matilde Vicenzi (\$14.99), with \emph{flavor: amaretto d'italia macaroons} selected &
Seven cookie products lack the \emph{quality ingredients} attribute; two gift baskets (\$100) exceed the price ceiling. \\
\addlinespace
``i am looking for some alcohol free skin care, and price lower than 80.00 dollars'' &
Gaelle Organic Exfoliant Sup\'erieure natural facial scrub (\$48.00); no options &
All nine are facial or body scrubs and serums under \$80, but none is described as \emph{alcohol free}. \\
\bottomrule
\end{tabular}
\end{table*}

\section{Example \webshop instructions}
\label{sec:supp_examples}

\Cref{tab:supp_examples} shows four \webshop tasks as the models saw them, abridged to the instruction, the correct product and the reason each of the nine distractors falls short. Every distractor is a similar product to the target and is taken from the same Amazon search. What disqualifies a distractor is a missing required attribute, an option (such as color, shape, size) that is not offered, or a price above the listed price ceiling. The models received the full title, price, features, description, option menus and image of all ten products (\cref{sec:supp_prompt}).

\section{Reproducibility}
\label{sec:supp_repro}

All tasks are frozen to JSON files with seeded construction, so every model saw identical candidate lists in identical order. Model identifiers, effort levels and the shared 16,000-token output budget are recorded in a single configuration file; each response is logged with its raw text, parsed choice, confidence, reasoning-token count and provider metadata. A single analysis script recomputes every statistic, figure and table in the paper and supplementary from the logged responses. The task files, logs and code will be released later upon publication.

\end{document}